\documentclass{article} 

\usepackage[accepted]{icml2025} 
\usepackage{times}

\usepackage{amsmath,amsfonts,bm}

\def\eqref#1{equation~\ref{#1}}

\def\1{\bm{1}}

\DeclareMathAlphabet{\mathsfit}{\encodingdefault}{\sfdefault}{m}{sl}
\SetMathAlphabet{\mathsfit}{bold}{\encodingdefault}{\sfdefault}{bx}{n}

\usepackage{wrapfig}

\usepackage{url}

\usepackage{xcolor}
\definecolor{google_yellow}{HTML}{ffbc32}
\definecolor{google_red}{HTML}{f4433c}
\definecolor{google_blue}{HTML}{2d85f0}
\definecolor{google_green}{HTML}{009925}
\definecolor{Gray}{gray}{0.9}
\definecolor{mygreen}{rgb}{0.0, 0.5, 0.0}
\definecolor{myred}{rgb}{0.8, 0.25, 0.33}
\definecolor{myblue}{rgb}{0.19, 0.55, 0.91}
\definecolor{uclablue}{rgb}{0.15, 0.45, 0.68}
\definecolor{boxgreen}{rgb}{0.02, 0.66, 0.02}
\definecolor{boxred}{rgb}{0.66, 0.1, 0.1}
\definecolor{boxblue}{rgb}{0.01, 0.01, 0.73}
\definecolor{mygray}{gray}{0.4}
\usepackage[pagebackref=true,breaklinks=true,colorlinks]{hyperref}
\definecolor{cite_color}{HTML}{2d85f0} 
\definecolor{link_color}{RGB}{0, 48, 143}
\hypersetup{colorlinks,linkcolor={link_color},citecolor={cite_color},urlcolor={red}}  
\usepackage[skip=3pt,font=small]{caption}
\usepackage[pagebackref=true,breaklinks=true,colorlinks]{hyperref}
\usepackage[capitalize,noabbrev]{cleveref}
\usepackage{listings}
\usepackage[utf8]{inputenc}
\usepackage{framed}
\usepackage{url}
\usepackage{booktabs}
\usepackage{multirow}
\usepackage{amsmath}
\usepackage{amssymb}
\usepackage{mathtools}
\usepackage{amsthm}
\usepackage{enumerate} 
\usepackage{microtype}
\usepackage{graphicx}
\usepackage{subfigure}
\usepackage{makecell}
\usepackage{pdfpages}
\usepackage{authblk}
\usepackage{scalerel}

\definecolor{xinyueblue}{rgb}{0.25, 0.44, 0.88}
\definecolor{xinyuered}{rgb}{0.79, 0.36, 0.27}

\definecolor{google_green}{HTML}{34A853} 
\definecolor{google_red}{HTML}{EA4335} 
\usepackage{pifont}

\def\titlelogo{\scaleobj{0.016}{\raisebox{-.22\height}{\includegraphics{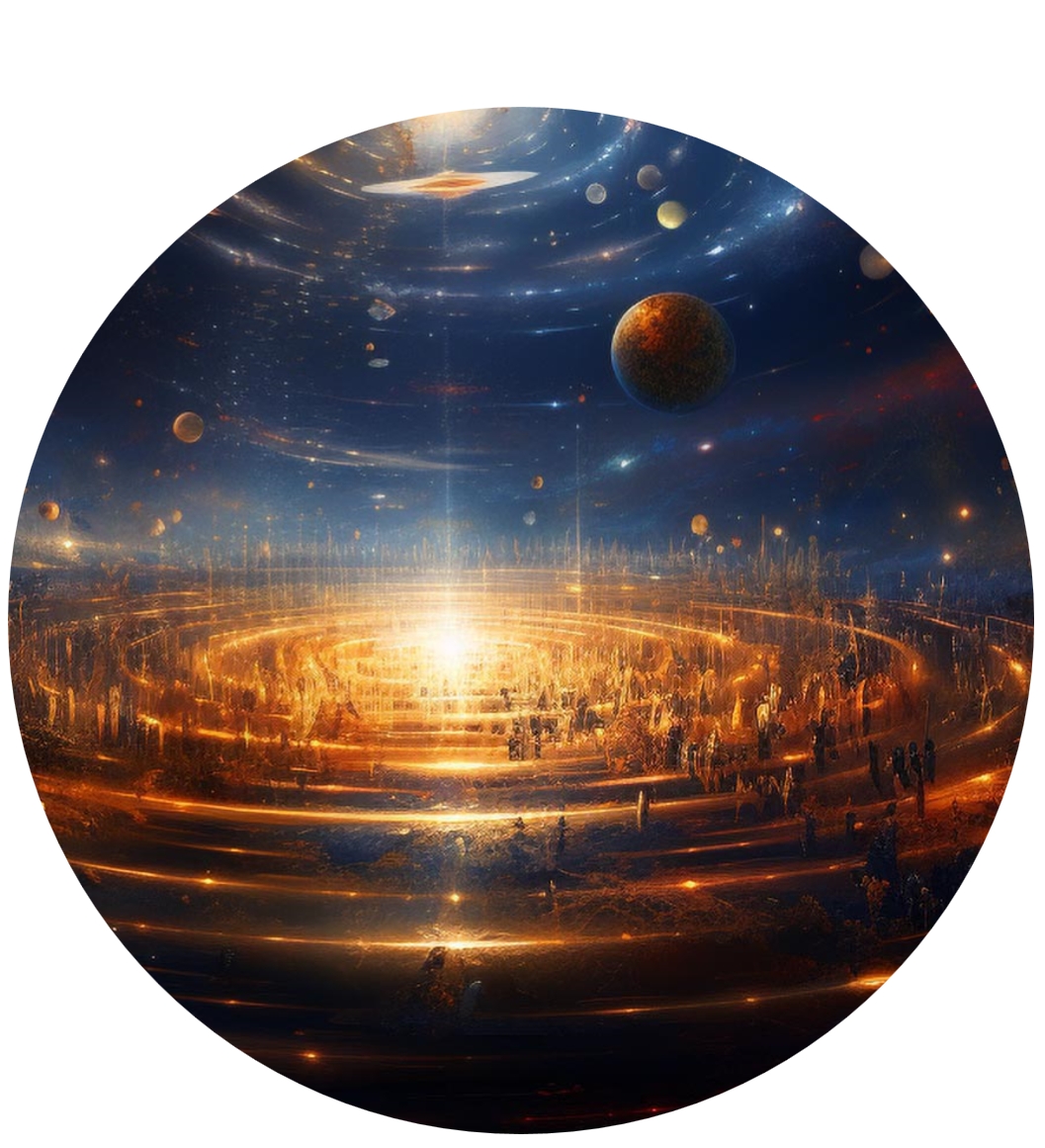}}}}

\icmltitlerunning{MCU: An Evaluation Framework for Open-Ended Game Agents}

\begin{document}
\thispagestyle{fancy}

\twocolumn[
\icmltitle{\begin{tabular}{ll}
\setlength{\tabcolsep}{0pt}
   \titlelogo &MCU: An Evaluation Framework for Open-Ended Game Agents \\
\end{tabular} }

\thispagestyle{firstpage}

\icmlsetsymbol{equal}{*}
\icmlsetsymbol{equal_co}{‡}

\begin{icmlauthorlist}
\icmlauthor{Xinyue Zheng}{bigai,equal}
\icmlauthor{Haowei Lin}{pku,equal}
\icmlauthor{Kaichen He}{pku}
\icmlauthor{Zihao Wang}{pku}
\icmlauthor{Qiang Fu}{tencent}
\icmlauthor{Haobo Fu}{tencent}
\icmlauthor{Zilong Zheng}{bigai,equal_co}
\icmlauthor{Yitao Liang}{pku,equal_co}
\end{icmlauthorlist}

\icmlaffiliation{bigai}{Beijing Institute for General Artificial Intelligence (BIGAI), Beijing, China}
\icmlaffiliation{pku}{Institute for Artificial Intelligence, Peking University, Beijing, China}
\icmlaffiliation{tencent}{Tencent AI Lab, Shenzhen, China} 
\icmlcorrespondingauthor{Zilong Zheng}{zlzheng@bigai.ai}
\icmlcorrespondingauthor{Yitao Liang}{yitaol@pku.edu.cn}

\icmlkeywords{Machine Learning, ICML}
\vskip 0.3in
]
\printAffiliationsAndNotice{* Equal contribution. ‡ Equal contribution as co-corresponding authors. All authors are affiliated with Team CraftJarvis.}

\begin{abstract}

Developing AI agents capable of interacting with open-world environments to solve diverse tasks is a compelling challenge. However, evaluating such open-ended agents remains difficult, with current benchmarks facing scalability limitations. To address this, we introduce \textit{Minecraft Universe} (MCU), a comprehensive evaluation framework set within the open-world video game Minecraft. MCU incorporates three key components: (1) an expanding collection of 3,452 composable atomic tasks that encompasses 11 major categories and 41 subcategories of challenges; (2) a task composition mechanism capable of generating infinite diverse tasks with varying difficulty; and (3) a general evaluation framework that achieves 91.5\% alignment with human ratings for open-ended task assessment. Empirical results reveal that even state-of-the-art foundation agents struggle with the increasing diversity and complexity of tasks. These findings highlight the necessity of MCU as a robust benchmark to drive progress in AI agent development within open-ended environments. Our evaluation code and scripts are available at \url{https://github.com/CraftJarvis/MCU}.

\end{abstract}

\section{Introduction}

\label{sec:intro}

Developing AI agents capable of interacting with dynamic environments—often referred to as ``open-world'' in the literature~\citep{open-world}—to solve diverse tasks remains a long-standing challenge in Artificial Intelligence~\citep{kejriwal2024challenges}. Among the various environments used to study AI agents, games have emerged as a prominent choice, as they provide real-world challenges within programmable simulators, offering valuable opportunities for real-world simulation~\citep{genie, gato, deepmind_sima}. Compared to other digital environments such as web/apps~\citep{web-env-autoagent, UI-TARS-ui, Showui-ui}, mobile platforms~\citep{mobile-pan, Mobile-agent}, and coding IDEs~\citep{code-challenge, swe-bench, agent-coder}, games present a higher degree of control complexity, similar to that encountered by physical robotic agents~\citep{robocasa, robo-gensim, robo-auto}. While compared to robotics, games support long-horizon planning tasks and enable safer and more efficient testing within sandbox environments. 


A crucial aspect of open-ended game agents is their \emph{generalizability}. The ultimate goal of developing AI agents is to deploy them in real, open-world environments, where they must solve tasks robustly across drastically different situations. With this in mind, early game agents have been studied in procedurally generated Atari-like environments with diverse configurations~\citep{procegen}. Recent efforts shift toward complex environments with greater freedom~\citep{minedojo,crafter,xland}, among which Minecraft stands out. Minecraft is a video game that provides procedurally generated open-world environments with a state space exceeding the number of atoms in the universe (we elaborate on the benefits of using Minecraft for open-ended game agents in~\cref{sec:minecraft_intro}). This vast environment allows agents to tackle infinite open-ended tasks using human-like actions in diverse situations. 

\begin{figure*}[t]
    \centering
    \includegraphics[width=0.9\linewidth]{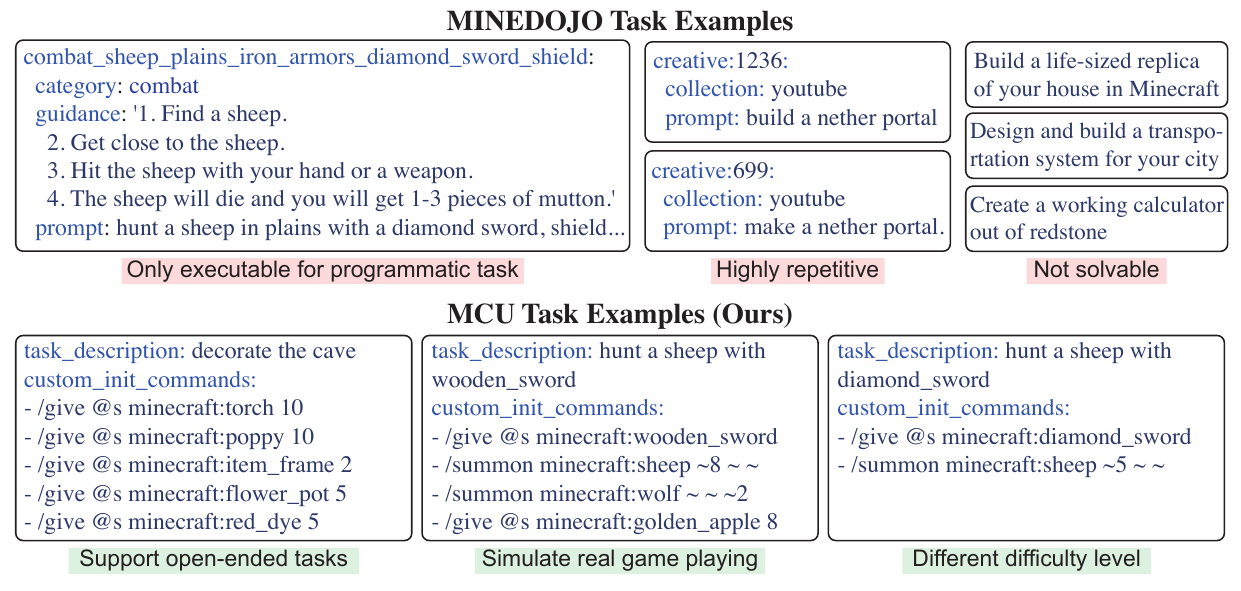}
    \caption{A comparison between the ``tasks'' in MCU and Minedojo~\citep{minedojo}. We investigate the task list provided by MineDojo and identify several issues. For example, only programmatic tasks that have clear reward signal can be executable in the benchmark; many tasks in their list are repetitive (both No.1236 and No.699 are ``build nether portal''); and a large amount of tasks in the creative tasks are not solvable even by human. To address this, our MCU benchmark can create executable configurations for open-ended tasks, and ensure intra-task and inter-task diversity to simulate real game playing in different difficulty levels, while preserving solvability of tasks.}
    \label{fig:mcu_vs_minedojo}
\end{figure*}

Despite the numerous advantages of Minecraft as an experimental environment, we have identified several limitations in existing benchmarks that impede a comprehensive evaluation of agents' generalizability. These limitations include low task quality~\citep{minedojo}, insufficient diversity~\citep{crafter}, and the lack of automatic evaluation suites~\citep{groot}. We further illustrate these issues in~\cref{fig:mcu_vs_minedojo}. Based on these issues, we find that Minecraft agents lack a unified benchmark, and each agent is evaluated on a distinct set of tasks (see~\cref{sec:related_work} for further discussion). To address this, we introduce \emph{Minecraft Universe} (\textbf{MCU}), an advanced evaluation framework in Minecraft (\cref{fig:framework}). MCU encompasses thousands of composable tasks and provides a scalable automatic evaluation system, designed based on the following key principles:
\begin{figure*}
    \centering
    \includegraphics[width=1\linewidth]{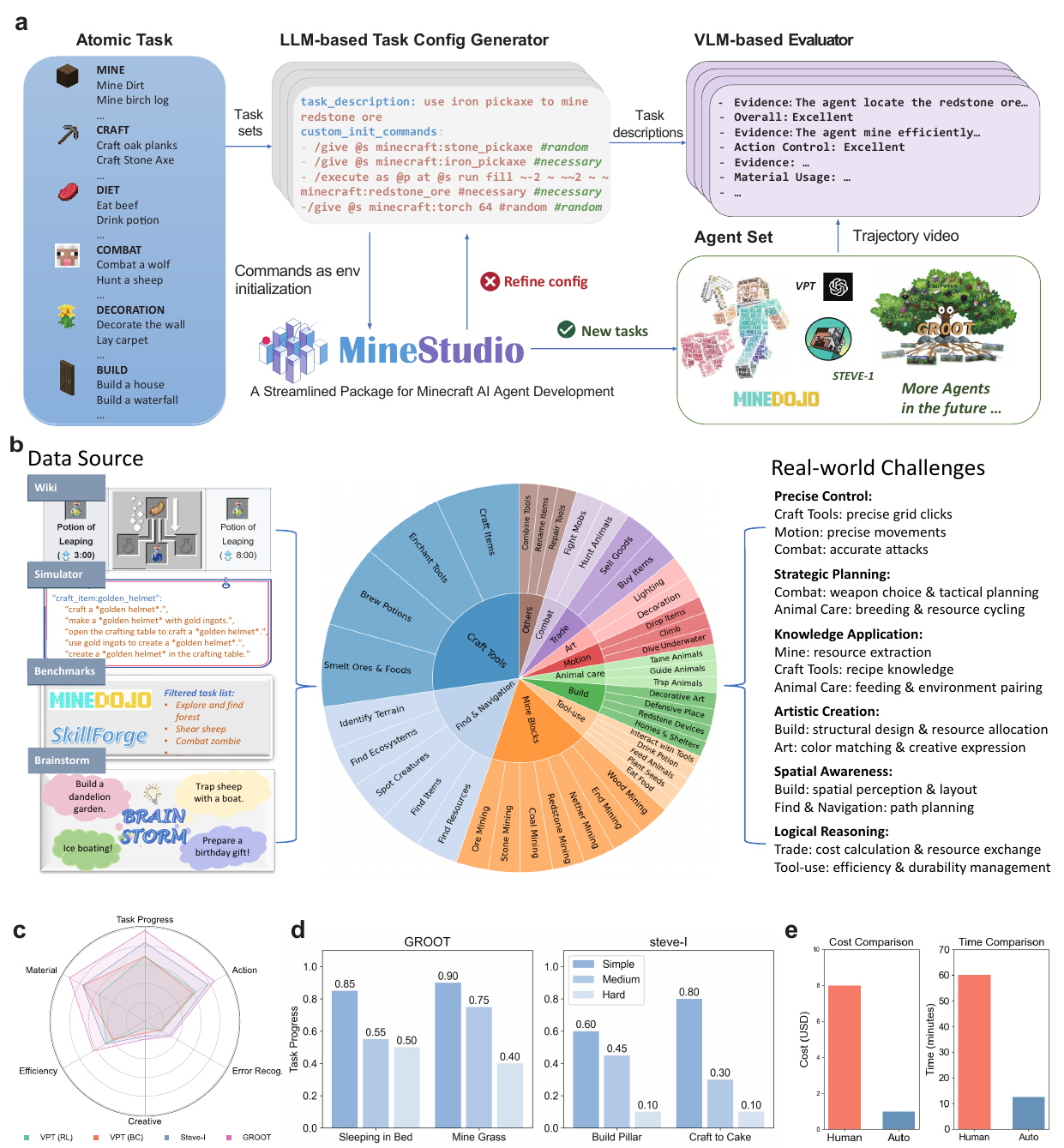}
    \caption{An overview of MCU. \emph{\textbf{a. Benchmarking pipeline.}} MCU includes two main components: \emph{\textbf{task generation} and \textbf{trajectory evaluation}}. The LLM-based task configuration generator instantiates the environment with the necessary prerequisites, random factors, and task descriptions for diverse atomic tasks. These configurations are verified using an environment simulator. The VLM-based evaluator assesses each task trajectory in video form across multiple dimensions, providing comprehensive performance insights. MCU offers a model-agnostic evaluation interface based on Minestudio~\citep{minestudio}, making it suitable for various agents. \emph{\textbf{b. Task category distribution.}} The atomic task set is sourced from the Minecraft wiki, in-game data, existing benchmarks, and brainstorming sessions. It spans 11 major categories and 41 subcategories, ensuring high \emph{\textbf{inter-task diversity}}. For readers unfamiliar with Minecraft, we illustrate the real-world challenges associated with different task categories to provide context. \emph{\textbf{c. Multi-dimensional capabilities.}} MCU evaluation shows that SOTA agents have made progress in overall task completion and material usage, but still have obvious limitations in creativity and error recognition. \emph{\textbf{d. Intra-task generalizability.}} Varying difficulty levels within the same task lead to performance degradation, which tests the agent's intra-task generalization capability. \emph{\textbf{e. Human vs AutoEval.}} The automatic evaluation (\emph{AutoEval}) of MCU is 8.1$\times$ more cost-effective and 4.8$\times$ more efficient in labeling 30 samples. Best viewed zoomed in.}
    \label{fig:framework}
\end{figure*}


\paragraph{High Task Diversity} 
Task diversity plays a crucial role in: (1) simulating challenges across various scenarios to thoroughly evaluate generalizability, and (2) harnessing the full potential of open-ended game agents capable of solving a wide range of tasks. MCU emphasizes two key aspects of task diversity in its implementation: (1) \textbf{Intra-task diversity:} We utilize a large language model (LLM) to dynamically generate task initialization conditions—such as biome, weather, and player states—introducing variability and randomness that closely mirror real gameplay scenarios (\cref{fig:framework}a). 
(2) \textbf{Inter-task diversity:} We collect 3,452 atomic tasks within Minecraft, spanning a broad spectrum of challenges, including precise control (\emph{e.g., combat, building}), logical reasoning (\emph{e.g., crafting, trading}), and knowledge application (\emph{e.g., tool use, animal care}) (\cref{fig:framework}b). These tasks can further be composed to generate new tasks on a combinatorial scale as detailed in~\cref{sec:task_suite}.

\paragraph{High Task Quality} Minecraft imposes numerous constraints that render unchecked tasks intractable~\citep{minedojo, taskgen_constraint}. For example, the task {``mine diamond with a wooden pickaxe”} is infeasible because diamonds cannot be mined with such a tool. Similarly, the task {``design and build a transportation system for your city”} from the MineDojo benchmark~\citep{minedojo} presents an extreme challenge even for humans. Other issues include repetitive tasks (\cref{fig:mcu_vs_minedojo}). To mitigate these issues, MCU filters tasks from a wide range of data sources, ensuring adherence to high-quality standards (\cref{fig:framework}b left). Additionally, in the LLM-based task configuration generator, we incorporate a series of soft constraints within the prompt, and implement a refinement mechanism based on self-reflection~\citep{reflexion} and the Minecraft simulator's feedback by executing the generated configuration. Further implementation details are provided in Section~\ref{sec:task_gen}.


\paragraph{Automated Evaluation} Open-ended tasks \citep{open-end, open-end2} inherently lack clear success signals and often depend on human evaluation or manually designed metrics~\citep{human-eval}, which hinders scalability. To address this issue, MCU introduces an automated evaluation system (\emph{AutoEval}) based on vision-language model (VLM) that fulfills two key objectives: (1) producing evaluation results that closely align with human judgments, and (2) offering multi-dimensional assessments that go beyond simple success rates for a comprehensive evaluation for open-ended tasks (\cref{fig:framework}c), while the designed ``task progress'' dimension is a process-supervised counterpart of 0-1 success rate. We also show in~\cref{fig:framework}e that \emph{AutoEval} is cost-efficient.

\paragraph{Enduring Challenges}
To ensure that MCU remains a long-term benchmark for agent development, we adopt two key strategies: (1) designing tasks with varying levels of difficulty, where increasing complexity introduces additional challenges such as adverse weather conditions and misleading factors. While state-of-the-art models achieve moderate success on simpler tasks, they struggle with more complex scenarios (\cref{fig:framework}d). (2) Enabling the composition of atomic tasks into more intricate tasks. This approach exponentially increases both the number and complexity of tasks, ensuring that MCU continues to provide a lasting challenge.

\section{\includegraphics[scale=0.01]{figs/mcu_logo.png} MCU Benchmark}

In this section, we first provide an overview of Minecraft and its game simulator, followed by a detailed outline of the construction process for the MCU benchmarking pipeline.

\subsection{Minecraft Environment}
\label{sec:minecraft_intro}


Minecraft is a voxel-based 3D video game that, due to its popularity and wide variety of mechanics, has become a prominent domain for RL research~\citep{rl-control,rl-deep}. Much of the prior work focuses on small, custom-built Minecraft environments with tasks such as navigation~\citep{navigation1,navigation2}, block placement~\citep{block1,block2}, combat~\citep{combat}, and other similar activities~\citep{activities}. More recent efforts have shifted towards studying the full, unmodified human action space, which encompasses tasks like drag-and-drop inventory management and item crafting. In this work, we employ unmodified Minecraft version 1.16.5 as our testing environment~\citep{minerl}, utilizing mouse and keyboard inputs as the action space and a 640 × 360 RGB image as the observation. The specifics of the action space will be detailed in~\cref{tab:actionspace}.


As mentioned in~\cref{sec:intro}, Minecraft serves as a powerful experimental environment due to its unparalleled diversity, complexity, and open-ended nature, which enable creative gameplay and countless possibilities. Below, we outline the key features that make Minecraft particularly suitable for open-ended game agent development: 

\begin{enumerate}
    \item \textbf{Vast State Space.} Minecraft provides an extraordinarily large state space, as illustrated in Table~\ref{tab:state_space_comparison} with an intuitive comparison. Its expansive maps, functional blocks, and diverse mobs result in an immense number of possible configurations. This makes Minecraft an ideal platform for studying the generalizability of agents. Additionally, we present illustrative experiments in Appendix~\ref{state_space} to demonstrate the importance of vast state spaces for agent generalization.

    \item \textbf{High Complexity Support.} Minecraft supports tasks requiring advanced problem-solving skills. For instance, \emph{obtain diamond} task~\citep{guss2019neurips} involves over 20 sub-goals, often takes hours to complete, and demands the ability to remember terrain and resource locations—showcasing the game's capability to facilitate complex task development and execution.

    \item \textbf{Open-Endedness.} Minecraft encourages unrestricted exploration, allowing players to engage in a wide spectrum of challenges. These range from defeating the ender dragon, which requires long-horizon decision-making~\citep{jin2023mini}, to building a house with precise control and creativity~\citep{zhang2020high}. This open-ended nature fosters the development of agents capable of handling diverse and dynamic objectives.
\end{enumerate}


\subsection{Integration with MineStudio}
\label{sec:minestudio}
To create a robust and user-friendly benchmark, we develop MCU based on MineStudio~\citep{minestudio}, an open-source software package designed specifically to facilitate agent development in Minecraft. MineStudio provides intuitive APIs, extensive documentation and tutorials, making it an ideal foundation for our benchmarking framework.

MineStudio offers users extensive customization options by allowing them to inherit from the \texttt{MinecraftCallback} class. This facilitates functionalities such as issuing cheat commands, logging episodes, and overriding observations. Leveraging this flexibility, our task configuration pipeline generates the necessary data for class \texttt{CommandsCallback}, \texttt{SummonMobsCallback} and \texttt{FastResetCallback} to initialize the environment. To ensure generality, MineStudio unifies the agent inference pipeline. Consequently, our evaluation pipeline observes only the generated trajectories, which are consistently formatted by \texttt{RecordCallback} for compatibility across diverse agents. Additionally, we employ \texttt{RewardsCallback} to support user-defined metrics, such as task success rates and our \emph{AutoEval} metrics, and to enable RL training with MCU evaluation results.





\begin{figure*}
    \centering
    \includegraphics[width=0.99\linewidth]{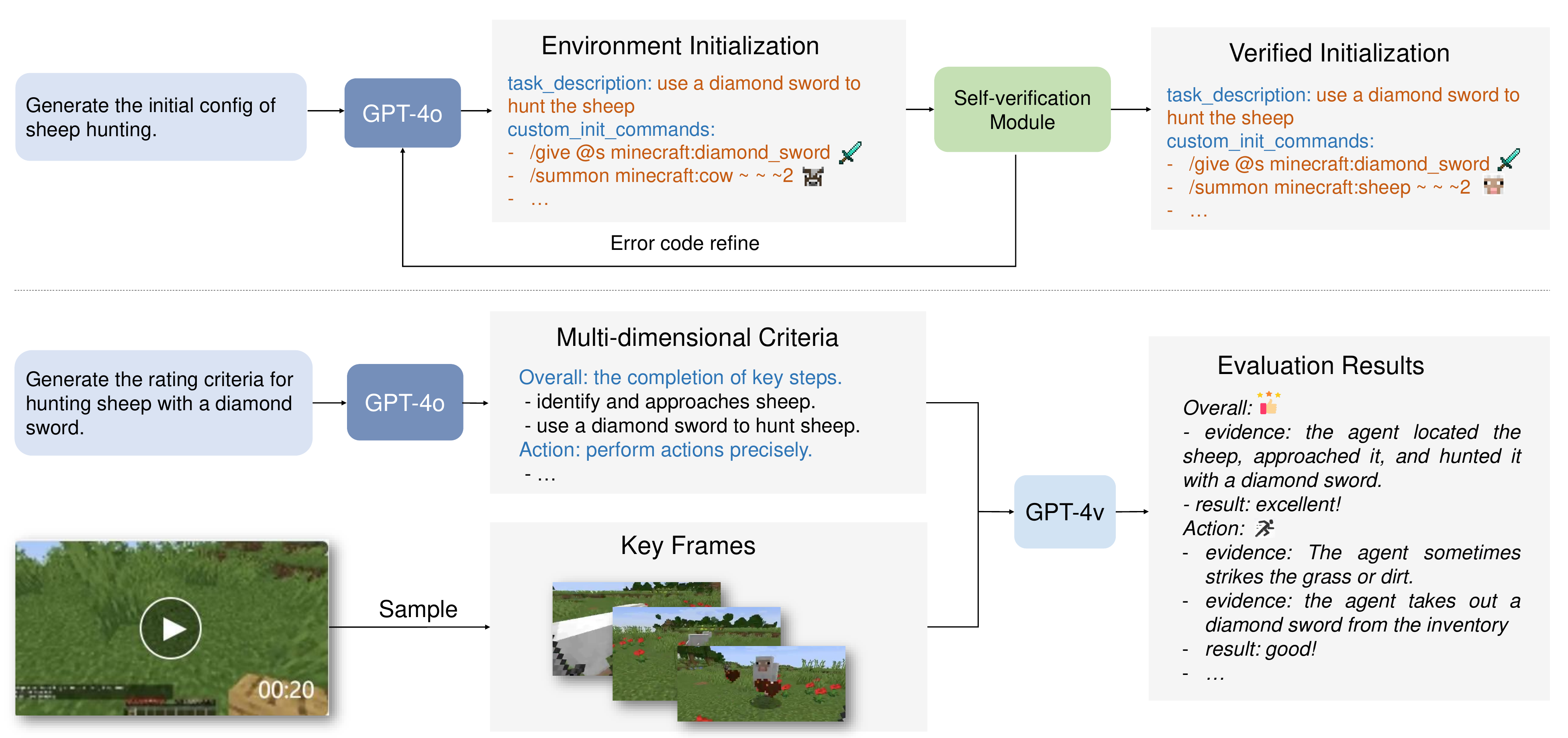}
    \caption{Automatic task generation and evaluation pipeline. \textbf{Top: Task Generation.} GPT-4o generates and verifies the environment initialization configuration for each atomic task, producing executable instructions compatible with Minestudio APIs. \textbf{Bottom: Automatic Evaluation.} A vision-language model (GPT-4o with vision) assesses task performance by analyzing sampled video frames against criteria generated by GPT-4o, providing detailed evidence and results. The GPT-4o used is \texttt{gpt-4o-2024-08-06}.}
    \label{fig:pipeline}
\end{figure*}


\subsection{Atomic Tasks: Fundamental Testing Units in MCU}
\label{sec:task_suite}
We introduce the concept of \textit{atomic tasks} as the fundamental testing units in MCU. The inspiration for atomic tasks stems from unit testing in software development~\citep{unit_test}, where a system is decomposed into smaller, independently verifiable units. Similarly, atomic tasks are designed to isolate and evaluate specific agent capabilities. 

An \textit{atomic task} $\mathcal{T}$ is defined solely by its goal $g$, independent of the methods, tools, or specific environmental conditions. During evaluation, an atomic task is instantiated, which induces a task-specific initial state distribution $\mathcal{P}(s_0|g)$ (see~\cref{sec:task_gen}). For example, the atomic task \texttt{mine stone} is purely goal-driven. Across different evaluation batches, it may manifest as \texttt{mine stone with a wooden pickaxe}, \texttt{mine stone during a rainy day}, or other variations. Regardless of these instances, the core goal remains consistent, ensuring a reliable test of whether the agent has genuinely acquired the underlying capability. This property makes atomic tasks an effective tool for evaluating an agent's ability to achieve goals under diverse and potentially unseen conditions.

Atomic tasks can also be combined to form more complex tasks using logical operators such as ``and'' ($\bigwedge$) and ``or'' ($\bigvee$), or by introducing constraints such as ``when,'' ``where,'' and ``how.'' For instance, an agent could be tasked with \texttt{[mine oak log] or [mine grass] \{bare-handed\} and \{then\} [craft sticks]}, where \texttt{[]} denotes individual atomic tasks. This compositional approach enables the exploration of a vast task space by leveraging the combination of simple goals to create more challenging scenarios. We also incorporate this feature in MCU task generation pipeline as detailed in~\cref{app:compose_pipeline}

To pursue task diversity, as outlined in~\cref{sec:intro}, we have collected \textit{3,452 atomic tasks}\footnote{The dataset is continuously expanding.}, each represented as a textual description. These tasks serve as the fundamental building blocks for task synthesis in MCU. They may be later composed (discussed in the previous paragraph) and instantiated using an LLM-based task configuration generator (see~\cref{sec:task_gen}). The atomic tasks are categorized into 11 major categories and 41 sub-categories, covering diverse challenges encountered in Minecraft. These tasks are sourced through the following pathways: (1) Distilling high-quality tasks from existing benchmarks, such as MineDojo~\citep{minedojo} and SkillForge~\citep{groot}. (2) Extracting tasks from the Minecraft Wiki\footnote{https://minecraft.wiki/}. (3) Synthesizing tasks using in-game data from the Minecraft simulator (e.g., generating tasks like \texttt{craft item X} if item ``X'' is craftable). (4) Brainstorming innovative tasks with input from domain experts and LLMs. Further details on the collection process for atomic tasks are provided in~\cref{sec:task-acquisition}.

\subsection{Task Configuration Generation}
\label{sec:task_gen}

To execute automatic or compositional tasks in the Minecraft environment, tasks must be \emph{configured} by specifying parameters such as spawn biomes, player states, inventories, and surrounding mobs. We formalize this configuration process as sampling an initial state $s_0$ from a task-specific distribution $\mathcal{P}(s_0 | \mathcal{T})$. To simulate realistic gameplay, evaluate agents effectively on a given task $\mathcal{T}$, and ensure intra-task diversity (\cref{sec:intro}), we propose a scalable task configuration pipeline powered by LLMs. This pipeline incorporates an automatic verification mechanism to ensure task validity based on two criteria: (1) the sampled initial state $s_0$ exhibits high diversity, and (2) the environment includes all elements necessary to complete the task (i.e., $s_0$ is compatible with $\mathcal{T}$). By leveraging the knowledge and creativity of LLMs, this approach generates diverse and scalable task scenarios, addressing the limitations of manually predefined configurations. The implementation details are as follows.

\paragraph{Config Generator}
As detailed in~\cref{sec:minestudio}, we utilize three callbacks from MineStudio to initialize the task environment. A powerful LLM GPT-4o~\citep{gpt4} is prompted with a task description and few-shot examples of the parameter format to populate these callbacks, generating executable configurations and detailed task descriptions for text-conditioned agents and evaluation in~\cref{sec:auto_eval}. For instance, the atomic task \texttt{mine diamonds} may include the description ``mine diamonds with an iron pickaxe", with the initial state $s_0$ providing necessary resources (e.g., diamond ores and an iron pickaxe) to eliminate preparatory subtasks. To enhance task diversity, random factors such as ore placement and item arrangements are introduced to prevent predictability. The prompt also includes soft constraints to address LLM limitations, such as numerical insensitivity or confusion with game-specific rules. For crafting tasks requiring precise materials (e.g., 3 wool blocks and 3 wooden planks), surplus resources are instructed to be generated to account for LLM inaccuracies. Soft constraints also ensure environmental integrity by avoiding the generation of inaccessible structures (e.g., via the \texttt{/fill} command). Complete prompt details are provided in Appendix~\ref{app:prompt_conf_gen}.

\paragraph{Verification}

To ensure the quality of the generated task configurations, the config generator incorporates a self-verification mechanism. Initially, the generated configuration is re-validated by the LLM using the Reflxion technique~\citep{reflexion}. If any errors are detected—such as summoning cows only for the task \texttt{hunt sheep} (\cref{fig:pipeline})—the config generator is prompted to regenerate the configuration. The generated configuration is also validated using the MineStudio Simulator to ensure executability. Error logs from the simulator are also utilized to guide further regeneration of the configuration if necessary.

\subsection{Automatic Evaluation (\emph{AutoEval})}
\label{sec:auto_eval}

Automatic evaluation is critical for machine learning benchmarks but challenging to design for AI agents. This problem is framed as defining a scoring function $f$ that maps a task description $\mathcal{T}$ and an agent’s trajectory $\textit{traj} = \{s_0, a_0, s_1, a_1, \ldots, s_T\}$ to a normalized score reflecting the trajectory's quality with respect to $\mathcal{T}$. In digital agent benchmarks, success rate based on annotated criteria is commonly used for tasks with clear, objective metrics (e.g., coding, OS operations). However, in open-ended games like Minecraft, defining a single programmatic metric is often infeasible. To address this, we propose a VLM-based multi-dimensional evaluation framework for MCU, comprising two components (\cref{fig:pipeline}): (1) criteria generation: creating clear, task-specific evaluation dimensions; and (2) scoring with criteria: leveraging predefined criteria to infer quality scores from agent performance videos using VLM.

\paragraph{Criteria Generation}
Preliminary experiments reveal that directly prompting GPT-4o (as a VLM) to score trajectories without task-specific criteria leads to suboptimal performance (\cref{tab:human_align_acc}). To address this, we introduce a criteria generation pipeline that provides detailed scoring instructions. Specifically, we define six key dimensions for evaluating agent performance in Minecraft. Prior to evaluation, GPT-4o is prompted to generate task-specific criteria for each dimension. The dimensions are as follows, with an example shown in~\cref{fig:pipeline} (bottom): (1) \textbf{Task Progress:} Assesses critical steps and factors required for task completion. (2) \textbf{Action Control:} Evaluates the avoidance of unrelated or unnecessary actions. (3) \textbf{Material Usage:} Measures the selection and application of materials. (4) \textbf{Task Efficiency:} Focuses on minimizing repetitions and optimizing strategies. (5) \textbf{Error Recognition:} Assesses the ability to identify and correct errors. (6) \textbf{Creative Attempts:} Recognizes innovative approaches in task execution. Please check~\cref{app:prompt_conf_gen} for the prompts of criteria generation.

\paragraph{Scoring with Criteria}
As described in~\cref{sec:minestudio}, the agent's rollout trajectories are recorded in video format. To balance resource efficiency and evaluative effectiveness under specific research conditions, we extract one frame every 30 frames from the video. During the evaluation phase, the sampled frames, along with task-specific criteria, are input into the VLM. The VLM assesses each dimension by identifying supporting evidence from the video, providing evidence and explanations, and subsequently assigning a score. For each criterion, we define five scoring intervals: \emph{very poor}, \emph{poor}, \emph{fair}, \emph{good}, and \emph{excellent}, corresponding to scores of 0, 0.25, 0.5, 0.75, and 1, respectively. Please check~\cref{app:prompt_single_rating} for the prompts of scoring.



\section{Experiments}

In this section, we first demonstrate the effectiveness of \emph{AutoEval}. Subsequently, we assess the capabilities of state-of-the-art agents using MCU and provide insights for the development of future open-ended Minecraft agents.

\subsection{Validity of Automatic Evaluation}

\paragraph{Experimental Setup} 
To validate the effectiveness of automatic evaluation, we compare the results of automatic evaluation methods with human annotations. We also evaluate a baseline approach, MineCLIP~\citep{minedojo}, which also focuses on automatic evaluation. Our evaluation is conducted under two distinct settings: (1) \textit{comparative evaluation}, where the quality of two trajectories of same task is compared; and (2) \textit{absolute rating}, where a score is assigned to a single trajectory. For \emph{AutoEval}, the comparative prediction for two trajectory is given by comparing the scores.

\paragraph{Dataset} We collected 500 trajectories spanning 60 tasks, derived from both agent rollouts and human gameplay videos. To evaluate the quality of these trajectories, we engaged 20 expert Minecraft players to provide annotations. The players were tasked with performing both comparative evaluations and absolute ratings on randomly sampled trajectory pairs for the same task or individual trajectories. Each player contributed 1 hours of work. Details regarding the annotation process are provided in~\cref{app:human_video}.

\begin{table}[t]
\centering
\caption{F1 scores for predicting the better human-annotated trajectory across different task categories. The compared methods include MineCLIP~\citep{minedojo}, our \emph{AutoEval} on both open-access models (MiniCPM~\citep{minicpm}, JarvisVLA~\citep{jarvisvla}) and closed API-based models (GPT-4o). The highest score for each task category is \textbf{bolded}.}
\Large
\resizebox{1\linewidth}{!}{
\begin{tabular}{@{}lccccccr@{}}
\toprule
\textbf{Method} & \textbf{Survive} & \textbf{Build} & \textbf{Craft} & \textbf{Mine} & \textbf{Explore} & \textbf{Average} \\ \midrule

MineCLIP    & 11.0            & 45.0          & 44.0                   & 73.0          & 0.0            & 34.6            \\
AutoEval (MiniCPM) & 65.0          & 43.0          & 80.0           & 59.0          & 53.0            & 60.0            \\
AutoEval (JarvisVLA) & 73.0          & 62.0          & 73.0           & 84.0          & 65.0            & 71.4            \\

AutoEval (GPT-4o)  & \textbf{100.0}  & \textbf{85.0}       & \textbf{62.0}                   & \textbf{73.0}          & \textbf{100.0} & \textbf{84.0}        \\ \bottomrule
\end{tabular}
}
\label{tab:human_align_acc}
\end{table}

\begin{table}[t]
\centering
\caption{F1 scores for predicting the better human-annotated trajectory across various dimensions (denoted using abbreviations).}
\Large
\resizebox{1\linewidth}{!}{
\begin{tabular}{@{}ccccccccr@{}}
\toprule
 \textbf{\makecell[c]{Progress}} & \textbf{\makecell[c]{Action}} & \textbf{\makecell[c]{Error Rec.}} & \textbf{\makecell[c]{Creative}} & \textbf{\makecell[c]{Efficiency}} & \textbf{\makecell[c]{Material}} & \textbf{\makecell[c]{Average}} \\ \midrule
 84.0 & 96.0 & 86.0 & 100.0 & 92.0 & 91.0 & \textbf{91.5} \\ \bottomrule
\end{tabular}
}
\label{tab:dimen_align}
\vspace{-5pt}
\end{table}


\paragraph{Comparative Evaluation} In this setting, participants are asked to vote on the comparative quality of the trajectory videos (denoted as \emph{A} and \emph{B}), selecting from the following options: \emph{A is better}, \emph{B is better}, \emph{tie}, or \emph{both are bad}. We filtered trajectory pairs annotated with first two options for automatic evaluation, resulting in 236 pairs. As shown in~\cref{tab:human_align_acc}, our methodology exhibits a significant improvement over MineCLIP, a CLIP model~\citep{clip} finetuned on large-scale Minecraft video frames. MineCLIP, however, struggles to capture complex event-level semantics with an average F1 score of only 34.6 (compared to 84.0 of \emph{AutoEval}). For intricate tasks such as \texttt{craft}, which demand precise attention to detail and the recognition of subtle elements, the performance of \emph{AutoEval} is slightly reduced. We leave further improvements in this for future work. Nevertheless, as demonstrated in~\cref{tab:dimen_align}, our automated evaluation metric achieves an average agreement rate of $91.5\%$ with human assessments across all dimensions. This highlights the effectiveness of the criteria-enhanced \emph{AutoEval} for multi-dimensional evaluations.

\paragraph{Absolute Rating}
For absolute rating, we collected a total of 227 individual ratings across approximately 50 distinct tasks. The Pearson and Kendall correlation coefficients between \emph{AutoEval} and human ratings are presented in~\cref{tab.absolute_rating}, demonstrating a strong overall positive correlation. However, the correlation varies across different dimensions. For instance, objective dimensions such as \emph{task progress} exhibit high agreement between human evaluators and VLM assessments, with a Pearson correlation of 0.78. In contrast, more subjective dimensions, such as \emph{creativity}, show a lower correlation. Additionally, we compute the inter-rater agreement for scoring the same trajectory, revealing a higher \emph{Pearson correlation} for task progress (0.83) and a lower correlation for creativity (0.69).

\begin{table}[t]
\centering
\caption{The Pearson correlation and Kendall's $\tau$ between \emph{AutoEval} and human ratings across different dimensions.}
\Large
\resizebox{0.99\linewidth}{!}{
\begin{tabular}{@{}lccccccc@{}}
\toprule
\textbf{Dimension} & \multicolumn{2}{c}{\textbf{Pearson}} & \multicolumn{2}{c}{\textbf{Kendall's $\tau$}} \\ \cmidrule{2-3} \cmidrule{4-5} 
 & Coefficient & P-value & Coefficient & P-value \\ \midrule
Task Progress & 0.78 & $1.70 \times 10^{-22}$ & 0.71 & $1.94 \times 10^{-19}$ \\
Action & 0.76 & $6.22 \times 10^{-19}$ & 0.67 & $1.68 \times 10^{-16}$ \\
Error Recog. & 0.68 & $1.10 \times 10^{-12}$ & 0.62 & $4.40 \times 10^{-10}$ \\
Creativity & 0.63 & $5.90 \times 10^{-9}$ & 0.56 & $1.18 \times 10^{-7}$ \\
Efficiency & 0.75 & $1.10 \times 10^{-16}$ & 0.67 & $1.02 \times 10^{-14}$ \\
Material & 0.70 & $2.28 \times 10^{-16}$ & 0.63 & $2.29 \times 10^{-13}$ \\ \bottomrule
\end{tabular}
}
\label{tab.absolute_rating}
\vspace{-7pt}
\end{table}

\begin{table*}[ht]
\vspace{-7pt}
\centering
\caption{Task progress for 35 tasks. Performance table across all tasks in simple mode. \textbf{Abbreviations:} 
\textbf{Exp} = exploring,\,
\textbf{FD} = find diamond,\,
\textbf{FF} = find forest,\,
\textbf{FV} = find villages,\,
\textbf{CM} = climb mountain;\,
\textbf{Crv} = carve,\,
\textbf{CO} = compose obsidian,\,
\textbf{Drk} = drink,\,
\textbf{F\&S} = flint\&steel,\,
\textbf{Slp} = sleep;\,
\textbf{Ck} = cake,\,
\textbf{Clk} = clock,\,
\textbf{CT} = craft table,\,
\textbf{DS} = diamond sword,\,
\textbf{Ld} = ladder;\,
\textbf{DO} = diamond ore,\,
\textbf{Dt} = dirt,\,
\textbf{Gr} = grass,\,
\textbf{IO} = iron ore,\,
\textbf{Ob} = obsidian;\,
\textbf{End} = enderman,\,
\textbf{Shp} = sheep,\,
\textbf{Skl} = skeletons,\,
\textbf{Spd} = spiders,\,
\textbf{Zmb} = zombies;\,
\textbf{BP} = build pillar,\,
\textbf{Cv} = cave,\,
\textbf{NP} = nether portal,\,
\textbf{SG} = snow golem;\,
\textbf{Wf} = waterfall;\,
\textbf{CTS} = crafting table from scratch,\,
\textbf{MDS} = mine diamond from scratch,\,
\textbf{D\&S} = dye and shear sheep,\,
\textbf{T\&P} = till and plant seeds,\,
\textbf{PAG} = prepare a gift. Best values within the same task are \textbf{bolded}.}
\label{tab:merged}

\Large
\begin{minipage}{\textwidth}
\centering
\resizebox{0.8\textwidth}{!}{%
\begin{tabular}{lcccccccccccccccccc}
\toprule
\multirow{2}{*}{\textbf{Agent}} 
& \multicolumn{6}{c}{\textbf{Navigation Task}} 
& \multicolumn{6}{c}{\textbf{Tool-use Task}} 
& \multicolumn{6}{c}{\textbf{Crafting Task}} \\
\cmidrule(lr){2-7} \cmidrule(lr){8-13} \cmidrule(lr){14-19}
& Exp & FD & FF & FV & CM & \textbf{Avg} 
& Crv & CO & Drk & F\&S & Slp & \textbf{Avg}
& Ck & Clk & CT & DS & Ld & \textbf{Avg} \\
\midrule
GROOT
& 0.90 & 0.56 & 0.75 & 0.60 & \textbf{0.60} & 0.72 
& 0.20 & \textbf{0.10} & \textbf{0.40} & 0.10 & \textbf{0.85} & \textbf{0.33}
& 0.35 & 0.60 & 0.45 & \textbf{0.75} & 0.25 & 0.48 \\

Steve-I
& \textbf{0.95} & 0.50 & \textbf{0.95} & \textbf{0.90} & 0.35 & 0.73
& \textbf{0.45} & 0.00 & 0.35 & \textbf{0.20} & 0.10 & 0.22
& \textbf{0.80} & \textbf{0.70} & 0.45 & 0.20 & \textbf{0.70} & \textbf{0.57} \\

VPT (BC)
& 0.90 & \textbf{0.65} & 0.87 & 0.75 & \textbf{0.60} & \textbf{0.75}
& 0.25 & 0.00 & \textbf{0.40} & 0.10 & 0.45 & 0.24
& 0.45 & 0.35 & 0.30 & 0.50 & 0.45 & 0.41 \\

VPT (RL)
& 0.70 & 0.58 & 0.55 & 0.50 & 0.35 & 0.54
& 0.15 & \textbf{0.10} & 0.35 & 0.15 & 0.25 & 0.20
& 0.70 & 0.62 & \textbf{0.50} & 0.30 & 0.62 & 0.55 \\
\bottomrule
\vspace{-7pt}
\end{tabular}}

\end{minipage}

\vspace{0.5em}

\begin{minipage}{\textwidth}
\centering
\resizebox{1.0\textwidth}{!}{%
\begin{tabular}{lcccccccccccccccccccccccc}
\toprule
\multirow{2}{*}{\textbf{Agent}}
& \multicolumn{6}{c}{\textbf{Mining Task}} 
& \multicolumn{6}{c}{\textbf{Combating Task}}
& \multicolumn{6}{c}{\textbf{Building Task}}
& \multicolumn{6}{c}{\textbf{Compositional Task}} \\
\cmidrule(lr){2-7} \cmidrule(lr){8-13} \cmidrule(lr){14-19} \cmidrule(lr){20-25}
& DO & Dt & Gr & IO & Ob & \textbf{Avg}
& End & Shp & Skl & Spd & Zmb & \textbf{Avg}
& BP & Cv & NP & SG & Wf & \textbf{Avg}
& CTS & MDS & D\&S & T\&P & PAG & \textbf{Avg} \\
\midrule
GROOT  
& \textbf{0.81} & 0.70 & 0.90 & 0.56 & \textbf{0.40} & \textbf{0.67}
& \textbf{0.30} & 0.50 & \textbf{0.56} & 0.50 & \textbf{0.75} & 0.53
& 0.40 & \textbf{0.20} & \textbf{0.45} & \textbf{0.65} & 0.15 & \textbf{0.38}
& 0.45 & \textbf{0.71} & 0.05 & \textbf{0.50} & 0.19 & 0.23 \\

Steve-I
& 0.35 & \textbf{0.85} & \textbf{0.95} & \textbf{0.20} & 0.35 & 0.54
& 0.05 & 0.30 & 0.40 & \textbf{0.75} & 0.42 & \textbf{0.54}
& \textbf{0.60} & 0.10 & 0.30 & 0.05 & 0.05 & 0.13
& 0.45 & 0.35 & \textbf{0.30} & 0.20 & \textbf{0.30} & 0.13 \\

VPT (BC)
& 0.30 & 0.30 & 0.50 & 0.15 & 0.38 & 0.33
& 0.25 & \textbf{0.55} & 0.55 & 0.35 & 0.50 & 0.36
& 0.00 & 0.02 & 0.35 & 0.00 & 0.20 & 0.11
& 0.30 & 0.30 & 0.10 & 0.00 & 0.25 & 0.13 \\

VPT (RL)
& 0.15 & 0.35 & 0.37 & 0.05 & 0.35 & 0.22
& 0.15 & 0.35 & 0.35 & 0.25 & 0.30 & 0.28
& 0.10 & 0.02 & 0.40 & 0.13 & \textbf{0.25} & 0.23
& \textbf{0.90} & 0.50 & 0.00 & 0.00 & \textbf{0.30} & \textbf{0.34} \\
\bottomrule
\end{tabular}
}
\end{minipage}
\label{tab.inter_task}
\vspace{-7pt}
\end{table*}

\begin{table}[t]
\centering
\caption{Performance changes of GROOT for selected tasks in simple and hard modes. Each result is averaged over 10 seeds.}
\Large
\resizebox{0.99\linewidth}{!}{
\begin{tabular}{lcccccc}
\toprule
\multirow{2}{*}{\textbf{Task}} & \multicolumn{3}{c}{\textbf{Task Progress}} & \multicolumn{3}{c}{\textbf{Action Control}} \\ 
\cmidrule(lr){2-4} \cmidrule(lr){5-7} 
 & \textbf{Simple} & \textbf{Hard} & $\Delta$ & \textbf{Simple} & \textbf{Hard} & $\Delta$ \\ 
\midrule
{Build Nether Portal}    & 0.45 & 0.30 & \textcolor{google_red}{-0.15} & 0.50 & 0.40 & \textcolor{google_red}{-0.10} \\
{Mine Iron Ore}   & 0.56 & 0.60 & 0.04 & 0.44 & 0.55 & 0.11 \\
{Craft to Cake}   & 0.35 & 0.32 & \textcolor{google_red}{-0.03} & 0.31 & 0.25 & \textcolor{google_red}{-0.06} \\
{Combat Skeletons} & 0.56 & 0.30 & \textcolor{google_red}{-0.26} & 0.25 & 0.20 & \textcolor{google_red}{-0.05} \\
{Carve Pumpkin}   & 0.20 & 0.15 & \textcolor{google_red}{-0.05} & 0.35 & 0.25 & \textcolor{google_red}{-0.10} \\
{Sleep in bed}    & 0.85 & 0.50 & \textcolor{google_red}{-0.35} & 0.40 & 0.30 & \textcolor{google_red}{-0.10} \\
\midrule
\textbf{Average}  & 0.50 & 0.36 & \textcolor{google_red}{-0.14} & 0.38 & 0.33 & \textcolor{google_red}{-0.05} \\
\bottomrule
\end{tabular}
}
\vspace{-7pt}
\label{tab:intra_task}
\end{table}

\subsection{Evaluating Existing Agents with MCU}
\label{sec:expr}

\paragraph{Agents} We evaluate four powerful foundation agents in Minecraft, all supported by MineStudio: (1) VPT (BC), a behavior cloning agent initially pre-trained on YouTube Minecraft videos and further fine-tuned on refined early-game contractor data; (2) VPT (RL), a reinforcement learning fine-tuned model based on VPT (BC) that maximizes the reward for obtaining diamonds in Minecraft; (3) STEVE-I~\citep{steve1}, which follows text instructions to complete tasks; and (4) GROOT~\citep{groot}, which solves tasks demonstrated by a reference video.  GROOT receives video instructions, STEVE-I receives text instructions, and VPT operates without instructions. We exclude agents that simplify the Minecraft environment by deviating from the native action space.
More details on these models can be found in~\citet{groot}.

\paragraph{Experimental Setup} 
While MCU enables scalable task evaluation without extensive human annotation, we carefully select a small yet diverse set of 30 atomic and 5 compositional tasks to illustrate our experimental conclusions without introducing excessive complexity. A comprehensive evaluation of a total of 150 tasks, including 90 randomly sampled atomic tasks and 60 compositional tasks, is deferred to~\cref{app:batch_inference}.
The 30 atomic tasks are drawn from six major categories, ensuring inter-task diversity while maintaining a moderate difficulty level suitable for Minecraft junior players. The difficulty of each task is of \emph{simple} mode, and each task is evaluated using 30 random seeds.
To highlight the importance of intra-task diversity, we select six atomic tasks (1-2 from each category), assessing both \emph{simple} and \emph{hard} modes of these tasks seperately. Each difficulty level is evaluated with 10 random seeds.

\subsubsection{Inter-task Generalization}
We present the \emph{AutoEval}-generated ``task progress'' scores for the inter-task generalization experiments in~\cref{tab.inter_task}, while the multi-dimensional performance, averaged over all tasks, is visualized in~\cref{fig:framework}c. 
Notably, the task progress metric is closely related to the commonly used ``task success rate'' in agent benchmarks. However, unlike success rate, which is a binary (0-1) outcome-based criterion, task progress is a process-supervised metric. This allows for a more fine-grained assessment of an agent's performance beyond simply determining whether a task was completed.

We observe that agents often struggle to reliably complete many tasks (e.g., \texttt{combat} and \texttt{build}). However, they still exhibit nonzero task progress, indicating partial progress as determined by the LLM. This suggests that while agents may not always achieve full task success, they are capable of making incremental advancements toward task completion.
Among the evaluated agents, we note that VPT (RL), which is specifically RL-tuned to maximize rewards for obtaining diamonds, performs well on tasks aligned with this objective (e.g., crafting a crafting table from scratch). However, it significantly underperforms on tasks unrelated to its target. This highlights the importance of inter-task diversity in assessing the generalizability of agents. While generalist agents such as GROOT and Steve-I demonstrate better open-endedness, they struggle with specific tasks (e.g., crafting particular items), indicating that future efforts should focus on improving performance in these areas. 
Additionally, we observe that task progress for compositional tasks is lower than for atomic tasks, underscoring the persistent challenge of MCU. Furthermore, as shown in~\cref{fig:framework}c, current agents fall short in creativity, error recognition, and efficiency, suggesting important directions for future research aimed at improving these aspects.

\subsubsection{Intra-task Generalization}

The task progress and action control performance of intra-task generalization experiments for GROOT, the best agent among the compared agents, are presented in~\cref{tab:intra_task}. The results indicate that for the same task, slight changes in the task situation lead to a significant drop in task progress performance. For example, consider the very simple task of \texttt{sleep in bed}, which only requires the agent to identify a bed and right-click the mouse. In \emph{simple} mode, the bed is placed directly in front of the agent on a plain with no surrounding objects. However, in \emph{hard} mode, both the bed and the agent are inside a house, requiring the agent to correctly identify the bed and interact with it. We observed failure cases where the agent mistakenly interacted with a chest in the room or left the house instead. This suggests that GROOT does not learn the skill robustly, a phenomenon not previously reported in the Minecraft agent literature.

Nevertheless, some tasks exhibited relatively stable performance across difficulty levels, such as \texttt{mine iron ore}, \texttt{craft to cake}, and \texttt{combat skeletons}. This may be because certain categories of tasks (e.g., building, tool-use, and crafting) are more susceptible to difficulty variations due to increased environmental complexity, while others are less affected. Overall, enhancing intra-task diversity of tasks is beneficial to test the robust generalzability of agents.

\section{Related Work}
\label{sec:related_work}
 
\paragraph{Benchmarks in Minecraft} 
MineDojo~\citep{minedojo} introduces a suite of 1,560 creative tasks defined by natural language instructions. However, it suffers from significant redundancy and overly complex tasks, complicating practical evaluation, as shown in~\cref{fig:mcu_vs_minedojo}. BEDD~\citep{milani2023bedd} defines five tasks covering various aspects of Minecraft, primarily designed for the MineRL BASALT competition. By decomposing the evaluation framework, BEDD facilitates detailed assessments of agent performance across subgoals and attributes such as human likeness. However, its reliance on human ratings limits scalability. Other works on Minecraft agents~\citep{deps,jarvis1,groot,plan4mc,vpt,steve1} lack a unified benchmark, with each agent evaluated on its own task set. We argue that establishing a standardized benchmark is a high priority for advancing Minecraft agent development.

\paragraph{Open-ended Agents in Minecraft} Many agents have been developed to interact with Minecraft environments.
Some methods focus on using Imitation Learning or Reinforcement Learning to learn various skills in open-world Minecraft to complete \textbf{short-horizon tasks}~\citep{vpt,groot,steve1,jiang2025reinforcement,zhao2024optimizing,groot15,yuan2024pre,cai2024groot,jiang2025visual}.
\citet{vpt} generates action labels from Minecraft videos on YouTube using IDM and performs imitation learning to obtain an unconditional policy capable of completing various tasks. \citet{steve1} uses MineCLIP as a text encoder to obtain a text-conditioned multitask policy based on unconditional VPT~\citep{vpt}.
Minecraft also supports the testing of \textbf{long-horizon programmatic tasks}, so some methods accomplish long-horizon tasks by using large language models as planners, combined with skill policies~\citep{deps,minedreamer,chen2024apt,plan4mc,mp5,zheng2023steve}.
\citet{deps} has designed an agent pipeline based on GPT, which interactively completes tasks from environmental feedback through the self-explanation and zero-shot planning capabilities of LLMs.
In order to further enhance the long-horizon capabilities of LLM Agents, some methods have explored efficient explicit memory mechanisms to support agents retrieve and improve from previous interaction trajectories~\citep{jarvis1,gitm,voyager,park2024mr,li2024optimus}.
Unlike designing complex agent pipelines, the rise of VLA~\citep{Brohan2023RT2VM,openvla} has inspired researchers to use end-to-end VLM as a policy to directly fulfill human instructions in Minecraft~\citep{omnijarvis,steve-2}.
Some methods use a code-as-policy approach~\citep{codeaspolicies}, using MineFlayer~\citep{mineflayer} as a language-conditioned policy, combined with a pretrained LLM to accomplish various tasks to avoid agents trapped on short-horizon skills lacking when executing long-horizon tasks~\citep{voyager,octupus,liu2024rl,liu2024odyssey}.
Finally, there are some methods focused on completing \textbf{open-ended creative tasks} in Minecraft, such as building and decoration, which differ from the traditional programmatic object-centric tasks and often cannot be directly defined by rule-based rewards~\citep{guo2024luban,zhang2023creative}.

\paragraph{LLM-as-a-Judge} 
The advancement of large language models (LLMs) has demonstrated remarkable performance in instruction following, query understanding, and response generation. This capability has motivated the use of LLMs as judges~\citep{llm-judge-mtbench}, leveraging their ability to score and rank model outputs. The strong performance of LLMs~\citep{brown2020language}, combined with well-designed assessment pipelines~\citep{li2023prd,beigi2024model,bai2024benchmarking}, enables fine-grained and detailed judgments for various evaluation applications, addressing the limitations of traditional evaluation methods that require extensive human annotation. Recent efforts have also explored the use of LLMs and vision-language models (VLMs) to evaluate AI agents~\citep{mobile-pan,zhuge2024agent}. Compared to previous work, MCU is the first to apply this paradigm to open-ended game agent task generation and evaluation. We argue that LLM-based task generators have the potential to create diverse tasks that are crucial for evaluating open-ended game agents, given the inherent open-endedness of LLMs~\citep{hughes2024open}. Furthermore, using the same LLM (or VLM) as a judge ensures more consistent evaluation criteria compared to crowdsourcing approaches~\citep{web-env-autoagent}.

\section{Conclusion}

We introduce \textit{Minecraft Universe} (MCU), a scalable evaluation framework for open-ended game agents in Minecraft. MCU features 3,452 diverse and composable atomic tasks, a dynamic task composition mechanism to ensure sustained challenges, and an automated evaluation system with over 90\% human alignment. Empirical results indicate that state-of-the-art agents struggle with tasks exhibiting high inter-task and intra-task diversity. To support standardized benchmarking, we release MCU-Turbo, a curated subset of 100 tasks with structured difficulty settings, as detailed in Appendix~\ref{app.mcu-turbo}. We hope MCU serve as robust foundations for advancing open-world agent research.

\section*{Acknowledgement}
This work was funded in part by the National Science and Technology Major Project (2022ZD0114902), and National Natural Science Foundation of China (62376031).
We acknowledge other members in CraftJarvis team for their data labeling support and constructive discussions, and thank Dr. Wenzheng Feng for his valuable insights. 

\section*{Impact Statement}
This paper presents research aimed at advancing the field of open-ended game agents, contributing to the broader understanding of adaptive and autonomous artificial intelligence in dynamic environments. While our work has the potential to influence various domains, including AI-driven decision-making and interactive entertainment, we do not identify any specific societal consequences that require immediate attention or emphasis at this time.

\bibliography{arxiv}
\bibliographystyle{icml2025}

\appendix
\onecolumn

\newpage

\lstset{
    language=Python,                     
    basicstyle=\ttfamily\small,          
    keywordstyle=\bfseries\color{blue},  
    stringstyle=\color{red},             
    commentstyle=\color{green!50!black}, 
    backgroundcolor=\color{gray!10},     
    showspaces=false,                    
    showtabs=false,                      
    frame=single,                        
    tabsize=4,                           
    breaklines=true,                     
    captionpos=b,                        
    numbers=left,                        
    numberstyle=\tiny\color{gray},       
}

\newpage
\section{Minecraft Environment Setting}
\label{app:minecraft_environment}
In the regular Minecraft game, the server (or "world") always runs at 20Hz while the client's rendering speed can typically reach 60-100Hz. To ensure consistency with the server, the frame rate is fixed at 20 fps for the client. The action and observation spaces in our environment are identical to what a human player can operate and observe on their device when playing the game. These details will be further explained in subsequent subsections. Additionally, diagnostic information such as in-game stats, contents of the agent's inventory, and whether any in-game GUI is open is provided by the environment. This information can only be used for tracking, recording, and evaluating purposes but cannot serve as inputs to evaluated agents.

\subsection{Minecraft Game World Setting}
We have chosen to conduct the test in Minecraft version 1.16.5's survival mode. During this open-world experiment, the agent may encounter situations that result in its death, such as being burned by lava or a campfire, getting killed by hostile mobs, or falling from great heights. When this happens, the agent will lose all its items and respawn at a random location near its initial spawn point within the same Minecraft world or at the last spot it attempted to sleep. Importantly, even after dying, the agent retains knowledge of its previous deaths and can adjust its actions accordingly since there is no masking of policy state upon respawn.

\begin{figure}[h]
    \centering\includegraphics[width=0.8\linewidth]{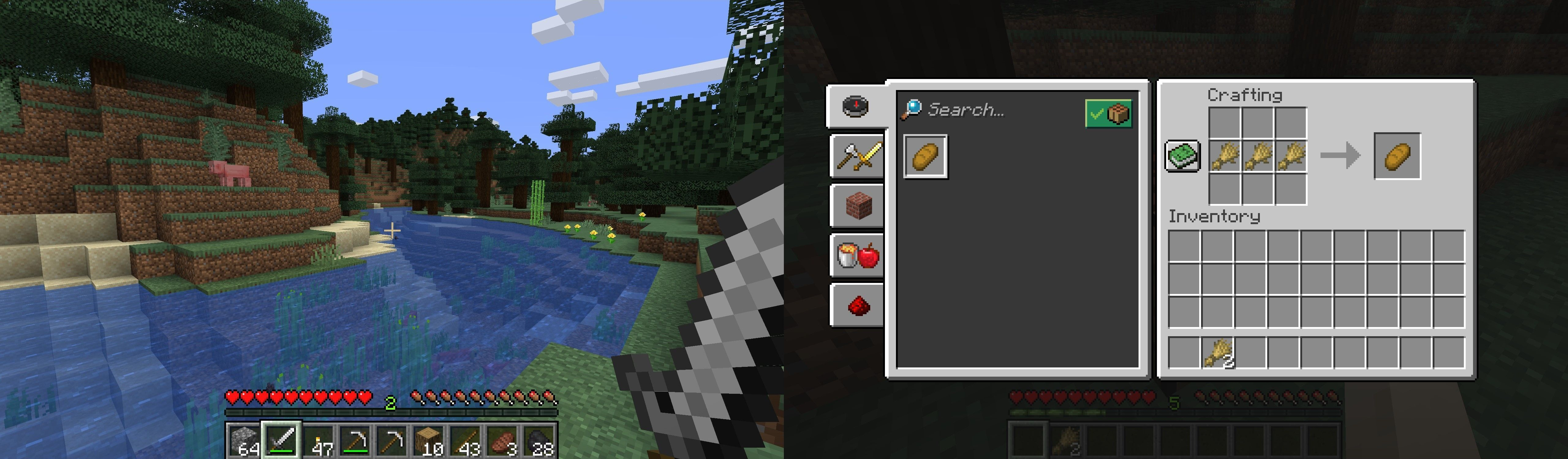}
    \caption{Minecraft game observation.}
    \label{fig:obs2}
\end{figure}

\subsection{Observation Space}
The observation space for a human player is limited to the raw pixels visible on the display screen. It does not include any hidden information from the game world, such as hidden blocks or nearby mobs. Additionally, any information contained in the pixels must be perceived by the model rather than directly given, including inventories and health indicators. Human players can access this information by pressing F3, which should be considered part of the game screen. There are no restrictions on optional parameters that human players can adjust in the display settings, such as field of view, GUI scale (controlling the size of in-game GUI), and brightness. The rendering resolution of Minecraft is 640x360; however, it is recommended to resize images to lower resolutions for better discernibility and computational efficiency.

\subsection{Action Space}
The action space is also consistent with human-playing settings, i.e., mouse and keyboard controls. These actions include key presses, mouse movements, and clicks. The specific binary actions that are triggered by keypress are shown in Table \Cref{tab:actionspace}.
In addition to actions triggered by keypresses, the action space also includes mouse movements. Similar to human gameplay, when there are no in-game GUIs open, moving the mouse along the X and Y axes changes the agent's yaw and pitch respectively. However, when a GUI is open, camera actions shift the position of the mouse cursor. The mouse movements are relative and adjust their position or camera angle based on their current state. More details can be found at \url{https://minecraft.fandom.com/wiki/Controls}

\begin{table}[h]
\centering
\renewcommand\arraystretch{1.2}
\caption{Binary actions included in the action space.}
\resizebox{0.99\linewidth}{!}{
\begin{tabular}{ccp{10cm}}
\hline
\textbf{Action} & \textbf{Human action} & \textbf{Description} \\
\hline
forward & W key & Move forward. \\
back & S key & Move backward. \\
left & A key & Strafe left. \\
right & D key & Strafe right. \\
jump & space key & Jump. \\
inventory & E key & Open or close inventory and the 2x2 crafting grid. \\
sneak & shift key & Move carefully in the current direction of motion. In the GUI it acts as a modifier key: when used with an attack it moves item from/to the inventory to/from the Hotbar, and when used with craft it crafts the maximum number of items possible instead of just 1. \\
sprint & ctrl key & Move fast in the current direction of motion. \\
attack & left mouse button & Attack; In GUI, pick up the stack of items or place the stack of items in a GUI cell; when used as a double click (attack - no attack - attack sequence), collect all items of the same kind present in inventory as a single stack. \\
use & right mouse button & Place the item currently held or use the block the player is looking at. In GUI, pick up the stack of items or place a single item from a stack held by the mouse. \\
drop & Q key & Drop a single item from the stack of items the player is currently holding. If the player presses ctrl-Q then it drops the entire stack. In the GUI, the same thing happens except for the item the mouse is hovering over. \\
hotbar.[1-9] & keys 1 – 9 & Switch active item to the one in a given hotbar cell. \\
show debug screen & F3 key & See the chunk cache, the memory usage, various parameters, the player's map coordinates, and a graph that measures the game's current frame rate. \\
\hline
\end{tabular}
}
\label{tab:actionspace}
\end{table}

\subsection{Why Minecraft is Suitable for Open-Ended Agent?}
\label{app:mc_challenge}

\begin{table}[h]
\centering
\caption{Comparison of State Space between Different Benchmarks. Unlike other benchmarks that isolate tasks within separate state spaces, which may simplify the learning process, Minecraft integrates all tasks into a shared state space. This requires the agent to generalize without relying on memorizing specific environments. 
The initial seed reflects the inherent generation capability of procedural generation, while the final state reflects the full range of possibilities contained within the entire game. The final state space of Minecraft (Approx. \(10^{10^{20}}\)) far exceeds the number of atoms in the universe (Approx. \(10^{80}\)). The constraint is $10^{-2}$, assuming that only one percent of the combinations are possible.}
\label{tab:state_space_comparison}
\resizebox{\linewidth}{!}{
\begin{tabular}{@{}lccc@{}}
\toprule
\textbf{Dimension}                & \textbf{Minecraft}                             & \textbf{Procgen/CoinRun}                     & \textbf{ALE/Pitfall}                  \\ \midrule
Initial Seed                      & $2^{64}$                                      & $2^{32}$                                    & Fixed                                 \\ \midrule
World Size                        & $30M \times 30M \times 384 \approx 3.46 \times 10^{18}$ & $64 \times 64$                              & Predefined Layout                     \\ \midrule
Block Types                       & 500                                           & 3                                           & 3-5                                   \\ \midrule
\textbf{Block States}             & \textbf{$500^{3.46 \times 10^{18}}$}           & \textbf{$3^{64 \times 64}$}                  & limited                                  \\ \midrule
Entities                          & \begin{tabular}[t]{@{}c@{}}
Mobs: 30+ Types, Health: 0--20 \\
Animals: 30+ Types, Health: 0--10 \\
Villagers: 13 Professions, 5 Levels, 100 Trades
\end{tabular}                     & Obstacles: 3 Classes                          & Obstacles: 8 Classes                        \\ \midrule
Entity Count                      & $10^7$                                       & $\leq 20$                                  & $\leq 10$                             \\ \midrule
\textbf{Entity States}                     & \begin{tabular}[t]{@{}c@{}}
$(30 \times 20 + 30 \times 10 + 
13 \times 5 \times 100)^{10^7}$ \\
$\approx 2.57 \times 10^{75 \times 10^7}$
\end{tabular}                     & $3^{20} \approx 3.49 \times 10^9$             & Limited                              \\ \midrule

\textbf{Inventory States}                       & \begin{tabular}[t]{@{}c@{}}
36 Slots, 500+ Item Types, Max Stack 64 \\
$ (500 \times 64)^{36}$
\end{tabular}                    & N/A                                           & N/A                                   \\ \midrule
\makecell[l]{\textbf{Final State Space} \\
\emph{Block $\times$ Entity $\times$  Inventory States $\times$ Constraint}} & 

\textbf{Approx. $10^{10^{20}}$}
                    & 
\textbf{Approx. $10^{99}$}
                  & Limited, Predefined Levels                   \\ \bottomrule
\end{tabular}
}
\end{table}

\subsubsection{State Space Calculation} 

Minecraft has an enormous state space. Here, we will estimate how many possibilities are contained within the full state space of Minecraft.

\paragraph{Block State}

The world in Minecraft consists of different types of blocks, each with potentially multiple states. Let \( B \) be the number of block types, and \( W \) be the world size, which is approximately \( 30M \times 30M \times 384 \), i.e., the total number of blocks in the world.

The formula for the number of block states is:

\[
\text{Block States} = B^{W}
\]

Substituting the values \( B = 500 \) and \( W \approx 3.46 \times 10^{18} \), we get:

\[
\text{Block States} = 500^{3.46 \times 10^{18}}
\]

In Procgen environments, the number of block states is estimated based on a pixel space. Block states are calculated based on how many different visual representations (pixels) can be generated. In ALE, the levels are artificially fixed and finite.

\paragraph{Entity State}

Minecraft contains a wide variety of entities including animals, mobs, and villagers, each with different properties and states. Mobs have more than 30 types, health ranging from 0 to 20, animals have more than 30 types with health from 0 to 10, and villagers have 13 professions, 5 levels, and 100 trades. The entire map contains approximately $10^7$ entities. Consider all possible combinations, the number of entity states is:

\[
\text{Entity States} \approx (30 \times 20 + 30 \times 10 + 13 \times 5 \times 100)^{10^7} \approx 2.57 \times 10^{75 \times 10^7}
\]

In Procgen environments like CoinRun, the number of entity states is typically smaller due to fewer types of entities and simpler interactions.

\paragraph{Inventory State}

Minecraft's inventory consists of 36 slots, each capable of holding a variety of items (over 500 types), with a maximum stack size of 64. The formula for the number of inventory states is:

\[
\text{Inventory States} = (500 \times 64)^{36}
\]

\paragraph{Final State Space}

The final state space is determined by the combination of block states, entity states, inventory states, and possible constraints (such as game rules). Assuming all these state spaces are independent, the final state space formula is:

\[
\text{Final State Space} = \text{Block States} \times \text{Entity States} \times \text{Inventory States} \times \text{Constraint}
\]

For Minecraft's final state space, assuming a constraint factor of \( 10^{-2} \), we get:

\[
\text{Final State Space} \approx 10^{10^{20}}
\]

\subsubsection{Further Attributes}

\paragraph{Complexity}
Minecraft presents a highly complex environment composed of diverse elements, including blocks, creatures, terrain, and vegetation. This complexity challenges agents to learn adaptive behaviors across varied tasks, fostering generalization in a dynamic setting.

\paragraph{Open-endedness}
The open-world nature of Minecraft exposes agents to a vast range of environments, requiring exploration and adaptive navigation. The flexibility to define tasks of varying difficulty enables targeted evaluation of agent capabilities across diverse challenges.

\paragraph{Dynamism and Unpredictability}
Unlike static benchmarks, Minecraft features dynamic environmental changes such as day-night cycles, emergent entities, and varied terrain. Agents must develop adaptability and robust decision-making to handle unforeseen events, enhancing their generalization to real-world complexities.

\paragraph{Creativity and Innovation}
Minecraft supports open-ended tasks like construction and decoration, encouraging agents to explore diverse strategies for goal achievement. This fosters innovation and problem-solving in complex, unstructured settings.

\paragraph{Broad Challenge Coverage}
Minecraft serves as an ideal platform for training and evaluating generalist agents, presenting four key challenges: \textbf{Long-horizon Decision Making:} Tasks decompose into flexible subtask sequences, requiring agents to plan beyond immediate actions. For example, acquiring wool may involve killing sheep, crafting from string, or trading with villagers, demanding strategic foresight. \textbf{Precise Control:} Building and crafting require fine-grained movement and accurate object manipulation. Tasks like constructing a Nether portal necessitate precise block placement, challenging agents to handle high-dimensional action spaces with stability. \textbf{Out-of-distribution Generalization:} The dynamic environment introduces novel scenarios beyond training data. Agents must generalize to unseen conditions, such as avoiding hazards (e.g., lava) or adapting to ecosystem variations. \textbf{Compositional Generalization:} Agents should infer new task compositions from learned subskills. For instance, if trained to craft sticks from planks and ladders from sticks, they should generalize to crafting ladders from planks. The vast combinatorial task space in Minecraft makes compositional generalization a crucial challenge.

\paragraph{Community and Resources}
Minecraft's extensive community provides rich datasets, strategies, and problem-solving techniques. Open-source mods and plugins further enable controlled experimental setups for agent training. 

\paragraph{Safe and Controlled Environment} Minecraft offers a risk-free virtual world where researchers can precisely manipulate environmental parameters for reinforcement learning, ensuring reproducibility and safety in training agents.

\newpage
\section{Details of Task Generation}

\subsection{The Source of Atomic Tasks}

\paragraph{Filtering from Existing Benchmarks} 
We curated tasks from established benchmarks, including Skill-Forge~\citep{groot}, MineDojo~\citep{minedojo}, and prior Minecraft research~\citep{vpt,plan4mc,deps,jarvis1}. To refine the selection, we performed deduplication to remove redundant tasks, excluded those too difficult for human players (e.g., constructing highly complex architectures or automated redstone circuits), and eliminated compositional tasks that could be decomposed into two or more atomic tasks.

\paragraph{Minecraft Wiki Resources} 
Additional tasks were sourced from the official Minecraft Wiki, which categorizes various in-game activities. From these, we extracted executable tasks, focusing primarily on the \texttt{Advancement} page\footnote{\url{https://minecraft.wiki/w/Advancement}}, which contains a curated list of diverse, engaging, and reasonably challenging tasks designed to enhance gameplay.

\paragraph{Synthesis from In-Game Information} 
The Minecraft simulator defines various item properties, such as ``craftable,'' ``mineable,'' ``eatable,'' and ``breakable.'' Leveraging these definitions, we systematically generated atomic tasks, such as \texttt{craft item X} if \texttt{X} is craftable, and similarly for other properties. This method efficiently scales up the task set while ensuring comprehensive coverage of game elements.

\paragraph{LLM and Expert Brainstorming} 
Beyond structured sources, we incorporated tasks generated through brainstorming sessions with expert Minecraft players and LLMs. This approach was particularly valuable for designing open-ended, creative tasks that pose real-world challenges for agents. Brainstorming was conducted in collaboration with a university Minecraft club.

Since most tasks are derived from the official Minecraft Wiki and in-game data, they are inherently reliable. Nevertheless, all tasks underwent rigorous validation through human inspection and automated scripts. The finalized atomic task list is provided as supplementary material alongside the code.

\label{sec:task-acquisition}

\subsection{Task Configuration}

In this section, we provide an overview of the key considerations for configuring a task, as introduced in~\cref{sec:task_gen}. This section aims to offer an intuitive understanding of task configuration. For detailed implementation and real-world configurations, please refer to Minestudio~\cref{sec:minestudio} and our integrated codebase.

The initial state of a task encapsulates all the information an agent can utilize based on its intended plan to execute the task. This includes not only the valid input but also any information the agent can derive or perceive, such as the observed 2D pixels of the game scene, inventory items, and coordinates (which can be accessed in-game by pressing F3, particularly the Y-dimension). The inventory $\mathcal{I}$ consists of two components: the necessary items for completing the task, denoted as $\mathcal{I}_n$—without which the agent would not be able to plan and execute the task in a real game—and additional random items, denoted as $\mathcal{I}_r$. Our objective is to manipulate these variables while ensuring that the random elements closely align with the real in-game distribution.

\subsubsection{Observation and Coordinate}

For a fixed version of the Minecraft game, the observation and coordinate elements are determined by the world seed, the coordinate, and the facing direction. The world seed is entirely independent of other variables and can be chosen arbitrarily. The facing direction remains unchanged from its state before teleportation to the task scene, making it inherently random and not subject to manipulation.

Setting a coordinate as a valid spawn location for a given task requires satisfying certain preconditions, such as specific biome types or other constraints defined by the game. For example, in the \texttt{climb the mountain} task, the agent must spawn in a \texttt{stony shore} biome, while for a task that involves reaching a village, the spawn location should be near one.

To facilitate reproducible task execution, we also collect a series of coordinate locations for each selected seed, corresponding to the required preconditions. Each (seed, precondition) pair can be mapped to multiple possible locations, allowing flexibility across different tasks. Minestudio allows for initializing game environment given these predefined random seeds.

\subsubsection{Inventories}

The inventory $\mathcal{I}$ consists of two main components: $\mathcal{I}_n$, the essential items required to complete the task, and $\mathcal{I}_r$, a set of random items acting as distractors. Since multiple approaches may exist to accomplish the same goal, $\mathcal{I}_n$ is also treated as a random variable. For instance, an agent may use either an iron pickaxe or a diamond pickaxe to mine diamond ore. To ensure comprehensive testing, we incorporate a variety of possible item sets for $\mathcal{I}_n$.

Regarding $\mathcal{I}_r$, we adjust its presence based on task difficulty. For some tasks, we omit $\mathcal{I}_r$ to reduce complexity. In other cases, we introduce random initial inventories by sampling from game snapshots derived from VPT contractor data. This approach ensures that the test environment remains diverse while maintaining a realistic distribution of inventory items.

\newpage

\section{Human Annotation}

\subsection{Minecraft Quiz}

To get an annotation for multi-dimensional task scores for trajectories used in our experiments (\cref{sec:expr}), we designed and distributed a questionnaire to confirm the annotators are familiar with Minecraft. The questionnaire is a quiz, containing five multiple-choice questions with 25 options to test their familiarity with Minecraft; each correctly answered option is worth 1 point. Then we filtered out the questionnaires with a correct rate of less than $75\%$, and then considered their investigation parts for the remaining questionnaires. The quiz is shown in~\Cref{tab:quiz}. We distributed the questionnaires to the signed up Minecraft annotators, and all of the annotators passed the quiz.


\begin{table}[h]
    \centering
    \caption{The quiz in our questionnaire (only 5 questions are presented), is used to judge the respondents' familiarity with Minecraft. The problems are adapted from~\citet{milani2023bedd}.}
    \label{tab:quiz}
    \resizebox{\linewidth}{!}{
    \renewcommand\arraystretch{1.1}
    \begin{tabular}{c|p{6cm}|l}
        \hline
        No. & Question & Options \\
        \hline
        1 & A bed can & \makecell[l]{
        A. speed up the night. \\
        B. change the respawn location. \\
        C. be crafted from drops of a certain animal in the game. \\
        D. can be crafted by a furnace, but cannot be crafted by a crafting table.
        } \\
        \hline
        2 & You can acquire EXP when & \makecell[l]{
        A. killing hostile mobs. \\
        B. mining trees. \\
        C. jumping on a coal ore block. \\
        D. mining coal. \\
        E. enchanting a diamond sword.
        } \\
        \hline
        3 & What mobs can deal damage to the player? & \makecell[l]{
        A. Skeletons. \\
        B. Zombies. \\
        C. Sheep. \\
        D. Pigs. \\
        E. Creepers. \\
        F. Enderman.
        } \\
        \hline
        4 & What items can be eaten? & \makecell[l]{
        A. Apples. \\
        B. Dirt. \\
        C. Beef. \\
        D. Wheat. \\
        E. Breads. \\
        F. Spider eyes.
        } \\
        \hline
        5 & \makecell[l]{If you mine a block with a bare hand, what \\ 
        kinds of block can drop the corresponding \\
        item?}  & \makecell[l]{
        A. Wooden logs. \\
        B. Wooden planks. \\
        C. Iron ore. \\
        D. Coal ore.
        } \\
        \hline
    \end{tabular}}
\end{table}

\subsection{Human Videos For Tasks}
\label{app:human_video}
Human videos serve two purposes: they are used as reference videos for GROOT and for comparison with the trajectory videos generated by the agent models. For each task, we select three world seeds: 19961103, 20010501, and 12345. For each (task, seed) pair, we manipulate the controllable parameters as described above, resulting in three distinct environment configuration files. For each configuration file, we record a corresponding human video. Additionally, we designate the first configuration file of seed 19961103 as the reference video for GROOT.

\subsection{Human rating system}

In our dataset, there are a total of 60 tasks, each containing 10 rollout trajectories in video format. These videos capture gameplay records from either humans or various agents. 

For the \textbf{absolute rating task}, we randomly select one task and present a corresponding video to human raters, who score it across six predefined dimensions. 

For the \textbf{comparative evaluation task}, we randomly sample two different rollouts from the same task, referred to as \textit{Video A} and \textit{Video B}. These videos may both be from human players, both from agents, or one from a human and the other from an agent. Raters are then asked to compare \textit{Video A} and \textit{Video B} on each dimension to determine which one performs better. 

\begin{figure}[H]
    \centering
    \includegraphics[width=\linewidth]{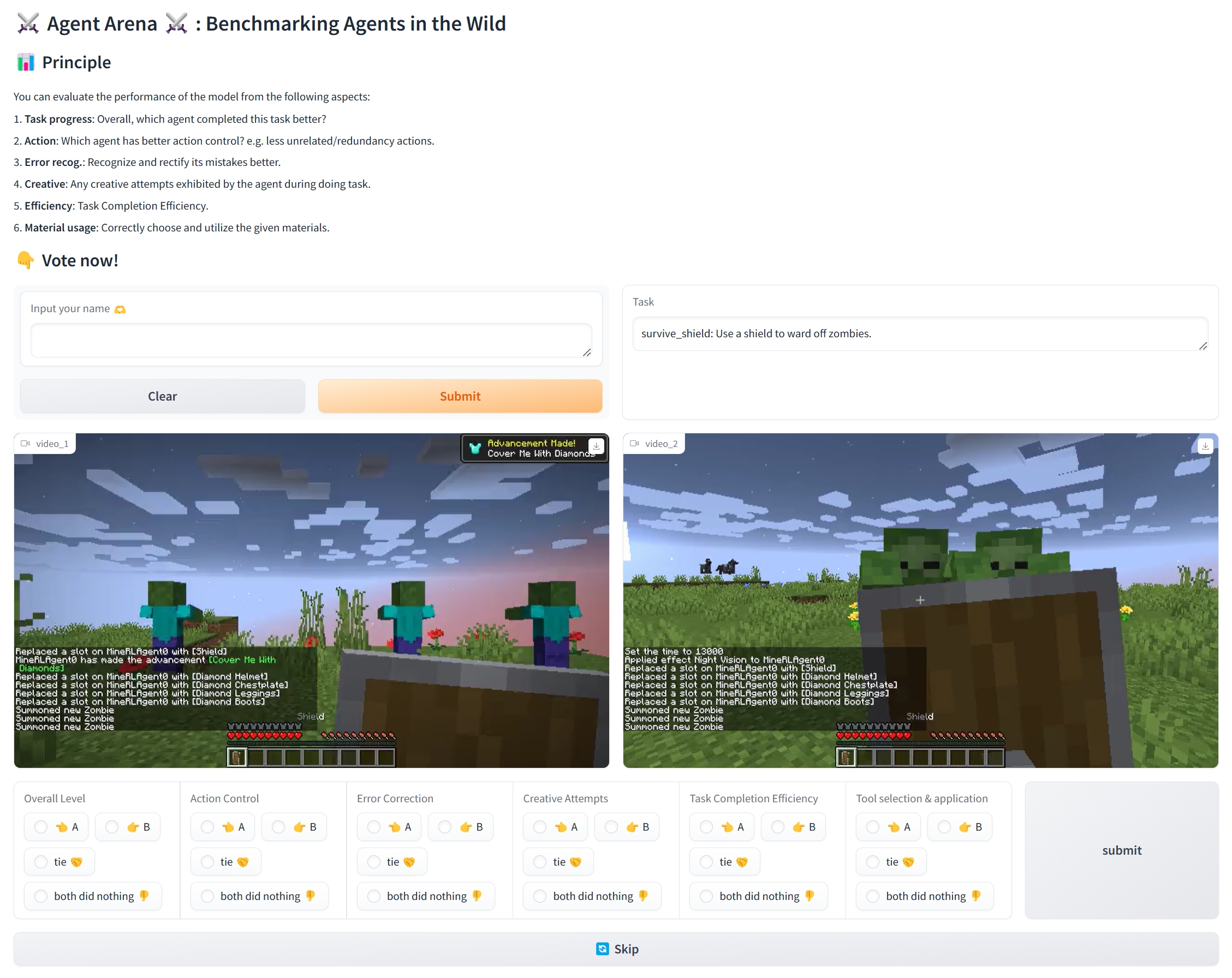}
        \caption{Video comparison website.}
        \label{fig:compare_web}
\end{figure}

\begin{figure}[H]
    \centering
\includegraphics[width=\linewidth]{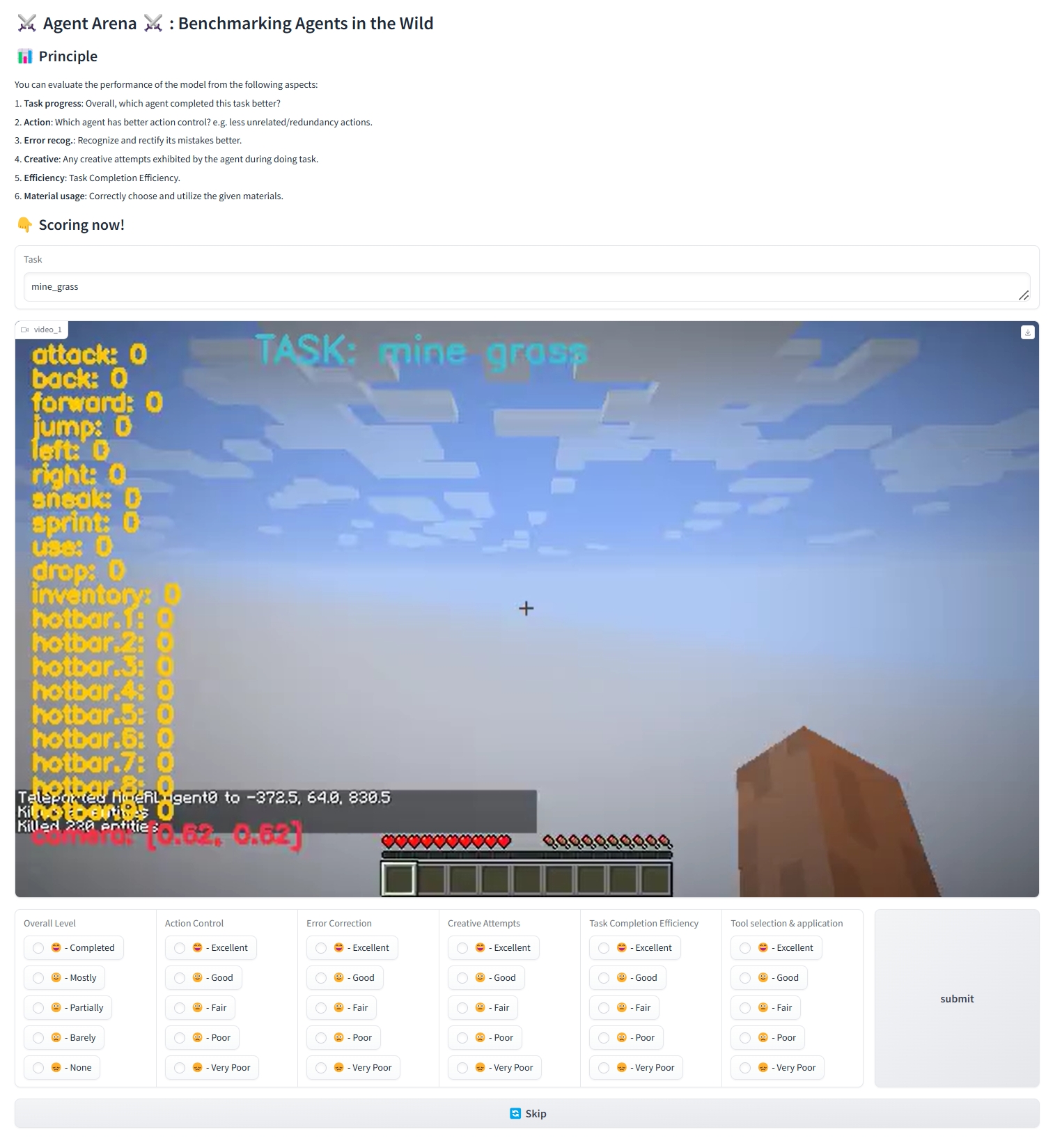}
    \caption{Individual video rating website.}
    \label{fig:single_web}
\end{figure}

The human rating interfaces are illustrated in Figure~\ref{fig:compare_web} and Figure~\ref{fig:single_web}. Taking the video comparison website as an example, it is designed to evaluate agent performance by displaying two videos side by side, enabling human raters to directly compare their behaviors within the same task. The interface consists of the following modules:

\begin{enumerate}
    \item \textbf{Task Description Module}: Positioned at the top-right, this module specifies the task to be evaluated (e.g., \textit{Survive Shield: Use a shield to ward off zombies}). It ensures that raters understand the objective before scoring.  

    \item \textbf{Video Display Module}: Two videos are presented side by side, each replaying an agent’s gameplay. This layout allows raters to observe agent behaviors, mistakes, or innovative strategies in real-time.  

    \item \textbf{Scoring Panel}: Located below the videos, this panel enables raters to assess agent performance across six dimensions. For each dimension, raters can indicate which agent performed better, mark a tie, or specify that neither agent took relevant actions.

    \item \textbf{Input and Submission Module}: At the top-center, an input box collects rater identifiers to ensure traceability. A \textit{Submit} button at the bottom sends completed ratings to the database, contributing to the dataset used for benchmarking.  
\end{enumerate}

\newpage

\section{Generalization Experiments}
\label{state_space}

In this experiment, we aim to demonstrate that developing open-ended agents requires selecting an environment with a vast state space. We seek to show that when the state space is limited, learning within such an environment is prone to overfitting rather than fostering genuine skill acquisition. Consequently, when tested in open-ended environments (i.e., the test set prepared for this experiment), the agent is likely to fail in handling unseen scenarios.

\begin{table}[H]
\centering
\caption{Training Hyperparameters}
\resizebox{0.6\linewidth}{!}{
\begin{tabular}{ll@{\hskip 2cm}ll}
\toprule
\textbf{Hyperparameter} & \textbf{Value} & \textbf{Hyperparameter} & \textbf{Value} \\ 
\midrule
Steps & 25M & GAE Lambda & 0.95 \\ 
Learning Rate & $2 \times 10^{-5}$ & PPO Clip & 0.1 \\ 
Scheduler & Linear & Policy Loss Weight & 1.0 \\ 
Optimizer & Adam & Value Loss Weight & 0.5 \\ 
Adam Epsilon & $1 \times 10^{-8}$ & KL Loss Weight & 0.3 \\ 
Number of Training GPUs & 2 & KL Loss Decay & 0.995 \\ 
Batch Size per GPU & 1 & Reward Discount & 0.999 \\ 
\bottomrule
\end{tabular}
}
\label{tab:hyper}
\end{table}

\begin{figure}[H]
    \centering
    \includegraphics[width=1.0\linewidth]{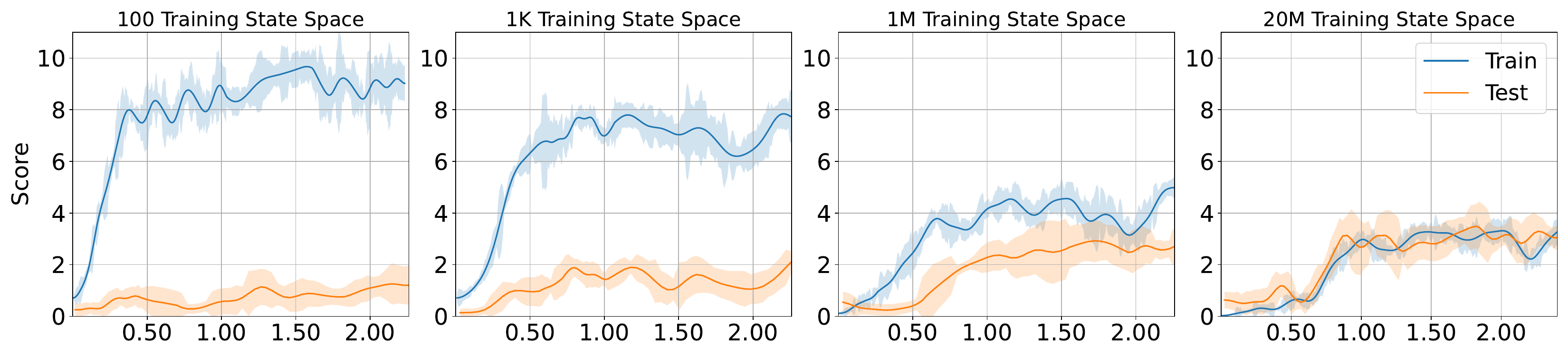}
    \caption{Generalization performance across different training state sizes. The x-axis values should be multiplied by $10^7$. We evaluate the agent on the full distribution of state space. The mean and standard deviation are computed over 100 episodes. As the number of training states increases, the gap between training and testing performance narrows: training curves become lower, while testing performance improves.}
    \label{fig:increase_state}
\end{figure}

\subsection{Experimental Setup} 
To assess the impact of state space size on generalization, we conduct experiments on the \emph{Hunt Sheep} task, using training sets ranging from $100$ to $10,000,000$ states. The state space primarily consists of variables such as equipment, mobs, biome, distance, and the number of sheep. For each setting, a subset of the state space is sampled as the training set, while evaluation is performed on the full state space. 

We train agents using online reinforcement learning with Proximal Policy Optimization (PPO) for 25 million steps, requiring approximately 50 training hours on three GPUs. Detailed hyperparameter configurations are listed in Table~\ref{tab:hyper}. The implementation is also supported by Minestudio.

\subsection{Results}
As shown in Figure~\ref{fig:increase_state}, agents exhibit strong overfitting when trained on small datasets. As the training set size increases, the generalization gap progressively narrows, with test performance improving as agents become more adept at generalization. To close the generalization gap, agents require exposure to as many as 10 million states.

However, we also observe certain limitations in the agent’s capacity. For example, at the start of the game, sheep may spawn behind the agent, outside its field of view. The agent fails to develop the behavior of scanning its surroundings before proceeding, often becoming distracted by other objects instead.

\subsection{Discussion}
These results highlight the necessity of complex environments like Minecraft, where achieving high performance in a limited state space does not necessarily imply strong generalization. Moreover, sufficiently large state spaces challenge existing reinforcement learning algorithms, offering insights into their capacity limits. At a critical state-space size, further increasing the diversity of states ceases to yield additional improvements in test performance, suggesting an upper bound on the agent’s generalization ability.

\newpage

\section{Large-Scale Inference and Evaluation of MCU}
\label{app:batch_inference}

This section demonstrates the MCU's capability for large-scale inference and task evaluation. We conduct an extensive experiment involving 150 tasks, consisting of 90 atomic tasks and 60 composite tasks. The experiments are performed on three agents: VPT (BC), VPT (RL), and STEVE-1. Due to the high cost of recording reference videos, GROOT is excluded from this evaluation. Specifically, atomic tasks are conducted in \emph{hard} mode, while composite tasks are executed in \emph{simple} mode.

\subsection{Atomic Tasks Setup}
For atomic tasks, one task is randomly sampled from the atomic task list for each experiment. As outlined in~\cref{sec:task_gen}, each selected task undergoes task configuration generation and verification to ensure the necessary preconditions for execution. This process guarantees that tasks are properly configured and validated before being executed in the experimental environment.

\subsection{Composite Tasks Setup}
\label{app:compose_pipeline}
Following the methodology described in~\cref{sec:task_suite}, we implement a composite task generation pipeline using logical connectors (``AND'' and ``OR'') to combine multiple atomic tasks. Composite tasks are designed in three distinct formats, illustrated by the following examples (note that users can define additional compositions):

\begin{enumerate}
    \item \textbf{Three Atomic Tasks Combined with ``AND'' or ``OR''}
    \begin{itemize}
        \item ``Find smooth red sandstone stairs OR mine yellow banner AND sell yellow dye''
        \item ``Find melon AND mine lodestone OR craft a wooden pickaxe''
    \end{itemize}
    
    \item \textbf{Two Atomic Tasks Combined with ``AND'' or ``OR''}
    \begin{itemize}
        \item ``Find melon AND mine lodestone''
        \item ``Craft a wooden sword OR find a diamond''
    \end{itemize}
    
    \item \textbf{Single Atomic Task with No Initial Tools Provided}
    \begin{itemize}
        \item ``Mine red sandstone from scratch''
        \item ``Craft a stone pickaxe from scratch''
    \end{itemize}
\end{enumerate}

In each case, atomic tasks are randomly sampled from a pool of over 3,000 available tasks. Composite tasks are designed to assess the system's ability to handle complex instructions and execute multiple tasks sequentially or in combination.

\begin{figure}
    \centering
    \includegraphics[width=0.95\linewidth]{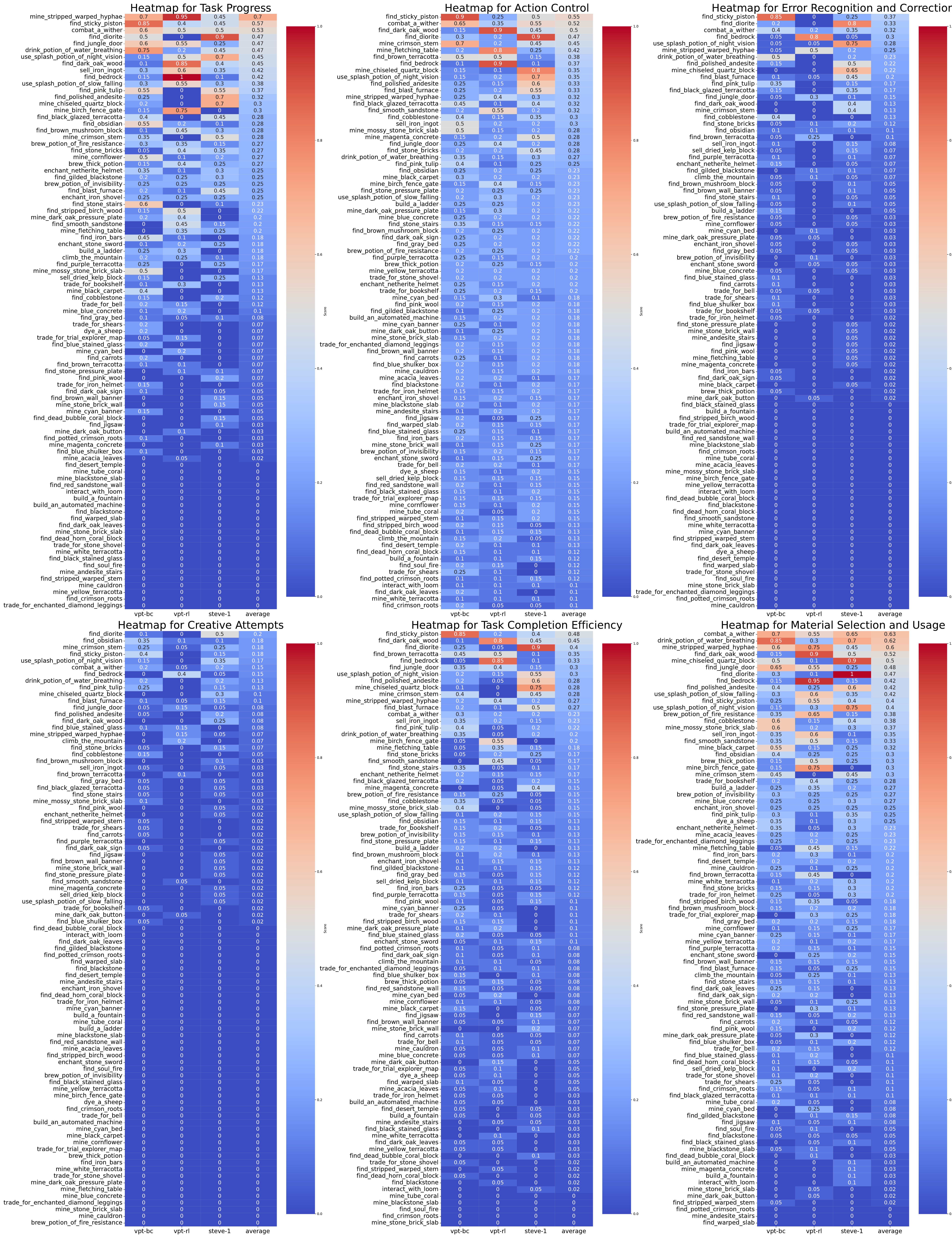}
    \caption{Performance of different agents across 90 atomic tasks.}
    \vspace{-3pt}
    \label{fig:atomic_res}
\end{figure}

\begin{figure}
    \centering
    \includegraphics[width=0.99\linewidth]{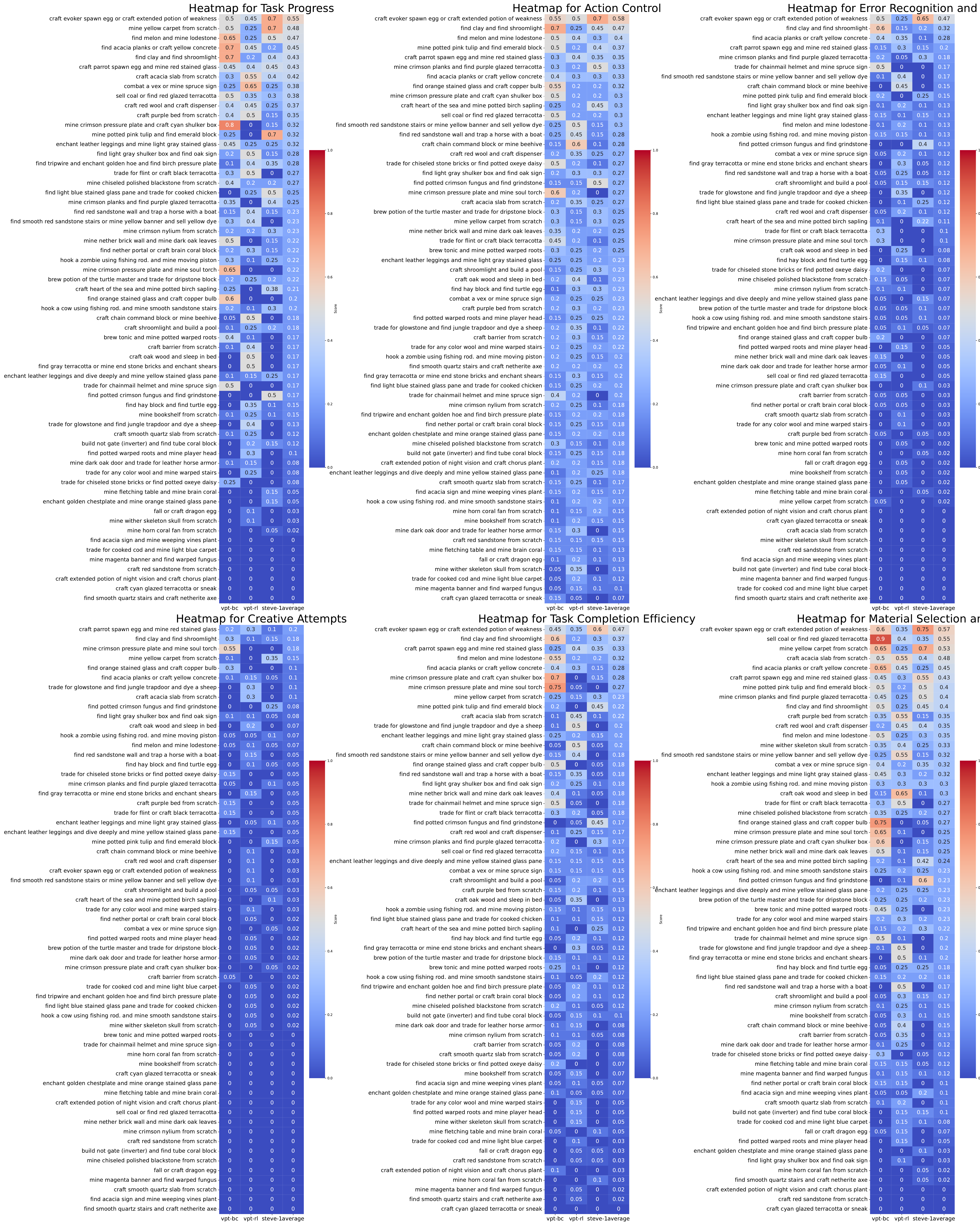}
    \caption{Performance of different agents across 60 compositional tasks.}
    \vspace{-3pt}
    \label{fig:compose_res}
\end{figure}

\subsection{Experimental Conclusions}
This experiment evaluates agent performance across a broad task domain, focusing on atomic tasks in \emph{hard} mode and compositional tasks. As shown in~\cref{fig:compose_res} and~\cref{fig:atomic_res}, agents struggle with these complex scenarios. STEVE-1 exhibits a performance decline in compositional tasks, likely due to its reliance on short prompts, as discussed in~\citet{steve1}. When encountering longer or unseen instructions, it experiences out-of-distribution (OOD) issues. The low performance of all agents in creativity and error recognition aligns with the conclusions drawn in~\cref{fig:framework}c.

\newpage

\section{MCU-Turbo: A Standard Benchmark for Evaluating Minecraft Agents}
\label{app.mcu-turbo}
We introduce MCU-Turbo, a canonical benchmark suite for systematically evaluating agents within the Minecraft Universal (MCU) framework. MCU-Turbo is designed as a standardized evaluation protocol, comprising 80 atomic tasks across 10 categories and 20 compositional tasks. Each task is assessed under two difficulty regimes—Simple and Hard—to rigorously test an agent’s capabilities in generalization, tool use, long-horizon planning, and robustness to environmental variation. The following experiments establish baseline performance using current state-of-the-art agents.

\subsection{Baseline Results in the Simple Evaluation Mode}
Agents demonstrate varied competencies across different evaluation dimensions under the simple setting. Notably, aspects such as creative behavior and error recognition remain challenging across the board, as illustrated in Table~\ref{tab:baseline_simple}.

\begin{table}[ht]
\centering
\caption{Baseline agent performance in simple mode.}
\label{tab:baseline_simple}
\resizebox{0.99\linewidth}{!}{
\begin{tabular}{lcccccc}
\hline
\textbf{Agent Name} & \textbf{Task Progress} & \textbf{Action Control} & \textbf{Error Recognition} & \textbf{Creative Attempts} & \textbf{Task Efficiency} & \textbf{Material Usage} \\
\hline
STEVE-1 & 31.4\% & 31.9\% & 13.1\% & 6.4\% & 23.2\% & 35.6\% \\
VPT (BC) & 29.1\% & 29.0\% & 11.8\% & 6.2\% & 21.3\% & 33.8\% \\
VPT (RL) & 25.9\% & 26.2\% & 8.8\% & 5.2\% & 18.9\% & 31.3\% \\
JARVIS-VLA & 25.6\% & 27.8\% & 9.3\% & 5.5\% & 18.3\% & 30.5\% \\
\hline
\end{tabular}
}
\end{table}

\subsection{Baseline Results in the Hard Evaluation Mode}
In the Hard mode, agents are subjected to increased environmental complexity and the presence of distractors. As shown in Table~\ref{tab:baseline_hard}, performance consistently declines across all dimensions, underscoring the difficulty of generalization under more challenging conditions.

\begin{table}[ht]
\centering
\caption{Performance degradation in hard mode.}
\label{tab:baseline_hard}
\resizebox{0.99\linewidth}{!}{
\begin{tabular}{lcccccc}
\hline
\textbf{Agent Name} & \textbf{Task Progress} & \textbf{Action Control} & \textbf{Error Recognition} & \textbf{Creative Attempts} & \textbf{Task Efficiency} & \textbf{Material Usage} \\
\hline
STEVE-1 & 23.1\% & 22.6\% & 6.9\% & 6.0\% & 16.6\% & 24.5\% \\
VPT (BC) & 23.0\% & 21.7\% & 6.2\% & 6.0\% & 15.6\% & 25.0\% \\
VPT (RL) & 21.0\% & 20.9\% & 5.0\% & 4.6\% & 14.4\% & 23.7\% \\
JARVIS-VLA & 20.9\% & 22.3\% & 5.5\% & 4.3\% & 14.5\% & 22.3\% \\
\hline
\end{tabular}
}
\end{table}

\subsection{Agent Performance on Creative vs. Programmatic Tasks}
Creative tasks present a substantially greater challenge compared to programmatic ones. For example, STEVE-1 exhibits a 15.8\% reduction in task progress when transitioning from programmatic to creative tasks, as shown in Table~\ref{tab:creative_vs_programmatic}. These results highlight the persistent difficulty of generalization in open-ended and less-structured settings.

\begin{table}[ht]
\centering
\caption{Performance comparison: creative vs. programmatic tasks.}
\label{tab:creative_vs_programmatic}
\resizebox{0.99\linewidth}{!}{
\begin{tabular}{lcccccc}
\hline
\textbf{Task Type} & \textbf{Task Progress} & \textbf{Action Control} & \textbf{Error Recognition} & \textbf{Creative Attempts} & \textbf{Task Efficiency} & \textbf{Material Usage} \\
\hline
Programmatic & 38.4\% & 36.5\% & 16.3\% & 7.7\% & 27.7\% & 43.8\% \\
Creative & 22.6\% & 26.2\% & 9.1\% & 4.7\% & 17.7\% & 25.4\% \\
\emph{Drop} & -15.8\% & -10.3\% & -7.2\% & -3.0\% & -10.0\% & -18.4\% \\
\hline
\end{tabular}
}
\end{table}

Overall, the MCU-Turbo benchmark provides fine-grained insights into agent capabilities across a diverse task spectrum. It emphasizes enduring challenges in creativity and adaptive behavior, and aims to steer future research toward the development of more general-purpose, robust Minecraft agents.

\definecolor{lightgray}{gray}{0.95}
\renewenvironment{shaded}{%
  \def\FrameCommand{\colorbox{lightgray}}%
  \MakeFramed {\advance\hsize-\width \FrameRestore}}%
 {\endMakeFramed}

\newpage  

\section{Prompts}
\UseRawInputEncoding
\subsection{Prompt for Config Generation of Atomic Tasks}
\label{app:prompt_conf_gen}
\begin{lstlisting}[language=, caption=Prompt for Config Generation of Atomic Tasks]
You are an expert of Minecraft, and I am a new Minecraft player.
You should give me all the necessary things I need for completing the task. 
I will give you the following information:

The task I want to complete: ...

You should perform the following steps to help me:
1. Tell me all valid items, mobs, biomes and all the necessary things to complete task;
2. Formulate the above information as cheat commands;
3. Randomly generate one or two related but not necessarily cheat commands.
4. Only output one simple task description, a thinking process and custom_init_commands.

e.g. The task I want to complete: Trade for iron helmet.
You should respond in the format as described below:
- In order to trade for iron helmet, we need at least 5 emerald and a armorer nearby.
- Task description: trade for iron helmet with a villager
- custom_init_commands:
  - /give @s minecraft:armor_stand 2
  - /give @s minecraft:emerald 10
  - /summon villager ~2 ~ ~-2 {Profession:"minecraft:armorer",VillagerData:{profession:"minecraft:armorer"}}
  - /give @s minecraft:diamond 64

e.g. The task I want to complete: craft a crafting table.
You should respond in the format as described below:
- In order to craft a crafting table, we need at least 4 planks.
- Task description: craft a crafting table
- custom_init_commands:
  - /give @s minecraft:oak_planks 64
  - /give @s minecraft:bread 16
  - /time set night

e.g. The task I want to complete: mine iron_ore.
You should respond in the format as described below:
- In order to mine iron_ore, we need at least a stone pickaxe or a better one, and have iron_ore nearby.
- Task description: mine iron ore with a stone pickaxe
- custom_init_commands:
  - /give @s minecraft:stone_pickaxe
  - /execute as @p at @s run fill ~2 ~2 ~3 ~1 ~5 ~4 coal_ore 
  - /execute as @p at @s run fill ~-5 ~-2 ~-1 ~ ~ ~-3 iron_ore 
  - /give @s minecraft:wooden_pickaxe

e.g. The task I want to complete: flying trident on a rainy day.
You should respond in the format as described below:
- In order to flying trident on a rainy day, we need a trident enchanted with the riptide enchantment, and set the weather in rainy mode.
- Task description: flying trident on a rainy day
- custom_init_commands:
  - /weather rain
  - /give @p minecraft:trident 
  - /give @p minecraft:trident{Enchantments:[{id:"minecraft:riptide",lvl:1}]} 3
  - /give @p minecraft:fire_charge{Enchantments:[{id:"minecraft:riptide",lvl:1}]} 3

e.g. The task I want to complete: combat a zombie.
You should respond in the format as described below:
- In order to combat a zombie, we need weapons, armors and a zombie nearby. Firstly, the Diamond Armor Set is the top-tier defensive gear, providing exceptional protection. Secondly, the Diamond Sword, can swiftly dispatch zombies. Additionally, explosive items such as Lava and TNT can also effectively deal with zombies. Zombies usually appear at night, so we need night vision.
- Task description: combat and kill a zombie
- custom_init_commands:
  - /replaceitem entity @s armor.head minecraft:diamond_helmet
  - /replaceitem entity @s armor.chest minecraft:diamond_chestplate
  - /replaceitem entity @s armor.legs minecraft:diamond_leggings
  - /replaceitem entity @s armor.feet minecraft:diamond_boots
  - /replaceitem entity @s weapon.mainhand minecraft:diamond_sword
  - /time set night
  - /effect give @a night_vision 99999 250 true
  - /summon minecraft:zombie ~3 ~ ~
  - /give @p minecraft:tnt 64

e.g. The task I want to complete: find a panda.
You should respond in the format as described below:
- In order to find a panda, we need to make sure there is a panda nearby.
- Task description: find a panda
- custom_init_commands:
  - /summon minecraft:panda ~ ~ ~3
  - /give @p minecraft:potato

e.g. The task I want to complete: interact with potion.
You should respond in the format as described below:
- In order to interact with a potion, you need at least one potion.
- Task description: interact with a potion
- custom_init_commands:
  - /weather rain
  - /give @s minecraft:potion 2

e.g. The task I want to complete: feed a sheep.
You should respond in the format as described below:
- In order to feed a sheep, you may need wheat in inventory and a sheep nearby.
- Task description: feed a sheep with wheat
- custom_init_commands:
  - /summon minecraft:sheep ~ ~ ~-2
  - /summon minecraft:sheep ~ ~ ~
  - /give @s minecraft:wheat 5

Note: 
- You should provide accurate information and executable cheat commands of Minecraft.
- The quantity of items in the cheat command should be more than what is required. For example, the task need at least 10 emerald, provide 20 instead. 
- You should provide all the tools and environments required for completing the task. 
- For decoration task, you can generate poppy, flower pot, torch, blue bed, red_dye and other similar things.
- Do not give me the final target things directly in my inventory.
- Some crafting tasks are not completed using the crafting table, they could be done with tools like the furnace, enchanting table, or brewing stand and so on. You need to select the appropriate tool.
- Remember to provide a crafting table, furnace, enchanting table, brewing stand or similar items, if the task requires it.
- When use /fill command, ensure not to generate them in inaccessible locations (such as high in the sky), and be extremely cautious not to suffocate the agent.
- The distance for summoning items should be within 4 blocks.
- For the "find" task, it is better to use /summon， /fill, or /execute
- Attention, there are certain items that cannot be directly summoned, such as trees, sugar cane, bubble_coral, etc. You should use /execute  or /give
\end{lstlisting}

\subsection{Prompt for Config Generation of Compositional Tasks}
\label{app:prompt_compose_gen}
\begin{lstlisting}[language=, caption=Prompt for Config Generation of Compositional Tasks]
You are an expert of Minecraft, and I am a new Minecraft player.  
You should give me all the necessary things I need for completing the task.  
I will give you the following information:

- The task I want to complete: ...

You should perform the following steps to help me:  
1. Tell me all valid items, mobs, biomes, and all the necessary things to complete the task.  
2. Formulate the above information as cheat commands.  
3. Only output one simple task description, a thinking process, and custom_init_commands.

e.g. The task I want to complete: Trade for iron helmet or mine stone.  
You should respond in the format as described as below:  
- In order to trade for iron helmet, we need at least 5 emeralds and an armorer nearby. In order to mine stone, we need a pickaxe, like a diamond pickaxe.  
- Task description: trade for an iron helmet with a villager and mine stone with a diamond pickaxe  
- custom_init_commands:  
    - /give @s minecraft:emerald 10  
    - /summon villager ~2 ~ ~-2 {Profession:"minecraft:armorer",VillagerData:{profession:"minecraft:armorer"}}  
    - /give @s minecraft:diamond_pickaxe 2

e.g. The task I want to complete: craft a crafting table and go explore.  
You should respond in the format as described as below:  
- In order to craft a crafting table, we need at least 4 planks. To explore, you need nothing.  
- Task description: craft a crafting table  
- custom_init_commands:  
    - /give @s minecraft:oak_planks 64

e.g. The task I want to complete: mine iron_ore and combat a zombie.  
You should respond in the format as described as below:  
- In order to mine iron_ore, we need at least a stone pickaxe or a better one, and have iron_ore nearby. In order to combat a zombie, we need weapons, armors, and a zombie nearby. Firstly, the Diamond Armor Set is the top-tier defensive gear, providing exceptional protection. Secondly, the Diamond Sword can swiftly dispatch zombies. Zombies usually appear at night, so we need night vision.  
- Task description: mine iron ore with a stone pickaxe and kill a zombie  
- custom_init_commands:  
    - /give @s minecraft:stone_pickaxe  
    - /execute as @p at @s run fill ~-5 ~-2 ~-1 ~ ~ ~-3 iron_ore  
    - /replaceitem entity @s armor.head minecraft:diamond_helmet  
    - /replaceitem entity @s armor.chest minecraft:diamond_chestplate  
    - /replaceitem entity @s armor.legs minecraft:diamond_leggings  
    - /replaceitem entity @s armor.feet minecraft:diamond_boots  
    - /replaceitem entity @s weapon.mainhand minecraft:diamond_sword  
    - /time set night  
    - /effect give @a night_vision 99999 250 true  
    - /summon minecraft:zombie ~3 ~ ~  

e.g. The task I want to complete: flying trident on a rainy day.  
You should respond in the format as described as below:  
- In order to fly with a trident on a rainy day, we need a trident enchanted with the riptide enchantment, and set the weather to rainy.  
- Task description: flying trident on a rainy day  
- custom_init_commands:  
    - /weather rain  
    - /give @p minecraft:trident{Enchantments:[{id:"minecraft:riptide",lvl:1}]} 3  

e.g. The task I want to complete: find bubble_coral and feed a sheep.  
You should respond in the format as described as below:  
- In order to find bubble_coral, we need to make sure there are bubble_corals nearby. In order to feed a sheep, you may need wheat in inventory and a sheep nearby.  
- Task description: find bubble_coral and feed a sheep with wheat  
- custom_init_commands:  
    - /execute as @p at @s run fill ~-5 ~-2 ~-1 ~ ~ ~-3 minecraft:bubble_coral  
    - /summon minecraft:sheep ~ ~ ~-2  
    - /summon minecraft:sheep ~ ~ ~  
    - /give @s minecraft:wheat 5

e.g. The task I want to complete: interact with a potion and eat bread.  
You should respond in the format as described as below:  
- In order to interact with a potion, you need at least one potion. In order to eat bread, you need at least one bread.  
- Task description: interact with a potion and eat bread  
- custom_init_commands:  
    - /give @s minecraft:potion 2  
    - /give @s minecraft:bread 2

---

Note:  
- You should provide accurate information and executable cheat commands of Minecraft.  
- The quantity of items in the cheat command should be more than what is required. For example, if the task needs at least 10 emeralds, provide 20 instead.  
- You should provide all the tools and environments required for completing the task.  
- Attention, there are certain items that cannot be directly summoned, such as trees, sugar cane, bubble_coral, etc. You should use /execute or /give.  
- For decoration tasks, you can generate poppies, flower pots, torches, blue beds, red_dye, and other similar things.  
- Do not give me the final target things directly in my inventory.  
- Some crafting tasks are not completed using the crafting table. They could be done with tools like the furnace, enchanting table, brewing stand, or similar tools. You need to select the appropriate tool.  
- Remember to provide a crafting table, furnace, enchanting table, brewing stand, or similar items if the task requires it. Use /give.  
- When using /fill command, ensure not to generate them in inaccessible locations (such as high in the sky), and be extremely cautious not to suffocate the agent.  
- The distance for summoning items should be within 4 blocks.
\end{lstlisting}

\subsection{Prompt for Criteria Generation}
\begin{lstlisting}[language=, caption=Prompt for Criteria Generation]
You are an expert of Minecraft and good at training agents in the AI field.  
I will give you a task description in Minecraft, and you need to generate the score points for assessing the completion of the task.

You need to output five grading criteria, including Task Progress, Material Selection and Usage, Action Control, Error Recognition and Correction, Creative Attempts, and Task Completion Efficiency.  
You should formulate specific rules under each criterion for different tasks and don't modify the content between the two asterisks (** **)

Building tasks should focus on whether the agent has completed the basic shape and structure.  
For example, the task name is "build a house and decorate the tree", please generate the score points for it.  
You should respond in the format as described below:

For build a house:
**Task Progress: the key factors/steps for completing the task**  
 - whether the agent builds four walls  
 - whether the agent builds a roof  
 - whether the agent builds a door  

**Action Control: whether the agents have unrelated operations of the task, including useless actions and redundant actions**  

**Error Recognition and Correction: whether the agent can promptly identify and rectify its mistakes**  
 - e.g., whether agents recognize the misaligned walls or incorrect material usage  
 - whether the corrected results demonstrate improvement and reduce flaws in the final product.  

**Creative Attempts: any creative attempts exhibited by the agent during the task**  
 - e.g., uniquely shaped rooms, distinctive decorative elements like furniture  

**Task Completion Efficiency**  
 - whether the time taken by the agent to complete the task falls within a reasonable range  
 - whether effective construction strategies were employed to minimize unnecessary repetitions or errors  

**Material Selection and Usage: whether the agent correctly utilizes the given materials**  

For decorate a tree:
**Task Progress: the key factors/steps for completing the task**  
 - Is there a tree in the image?  
 - whether the agent put something on the tree  

**Action Control: whether the agents have unrelated operations of the task, including useless actions and redundant actions**  
 - e.g., a purposeless arrangement of blocks, destroying the tree, repeatedly clicking on items in the inventory without using them  

**Error Recognition and Correction: whether the agent can promptly identify and rectify its mistakes**  
 - whether the corrected results demonstrate improvement and reduce flaws in the final product  

**Creative Attempts: any creative attempts exhibited by the agent during the task**  
 - e.g., Evaluate the overall visual effect of the decoration, including color coordination, layout rationality, and symmetry  
 - e.g., Are the decorations on the tree diverse and abundant?  

**Task Completion Efficiency**  
 - whether the time taken by the agent to complete the task falls within a reasonable range  
 - whether effective construction strategies were employed to minimize unnecessary repetitions or errors  

**Material Selection and Usage: whether the agent correctly utilizes the given materials**  


For example, the task name is "dig three holes and fill one", please generate the score points for it.  
You should respond in the format as described below:

**Task Progress: the key factors/steps for completing the task**  
 - whether the agent is digging the hole  
 - whether the agent digs three holes  
 - whether the agent fills one hole  

**Action Control: whether the agents have unrelated operations of the task, including useless actions and redundant actions**  
 - e.g., wandering aimlessly, destroying the tree  

**Error Recognition and Correction: whether the agent can promptly identify and rectify its mistakes**  
 - whether the corrected results demonstrate improvement and reduce flaws in the final product  

**Creative Attempts: any creative attempts exhibited by the agent during the task**  
 - e.g., using different tools to dig the holes like hands or pickaxes  

**Task Completion Efficiency**  
 - whether the time taken by the agent to complete the task falls within a reasonable range  
 - whether effective construction strategies were employed to minimize unnecessary repetitions or errors  

**Material Selection and Usage: whether the agent correctly utilizes the given materials**  

Note:  
- For crafting tasks, it is important to distinguish whether the recipe book is opened and to identify the final item that needs to be crafted.  
- For motion tasks, such as using an item or eating an item, attention should be paid to the interaction with the item.
\end{lstlisting}

\subsection{Prompt for Video Comparison}
\label{app:prompt_video_compare}
\begin{lstlisting}[language=, caption=Prompt for Video Comparison]
You are an expert in Minecraft and experienced in evaluating agents in the AI field.  
I will provide the following:

- A task name  
- Grading criteria for the task  
- Two videos (Video A and Video B) of an agent performing the task.

The grading criteria contain several major categories (surrounded by ** **) and several evaluation rules under each major category.  
You need to carefully compare the agent's performance in Videos A and B according to the evaluation rules and output one of the following:

- "A is better"  
- "B is better"  
- "tie"  
- "both are bad"

The more an agent complies with the rules in each criterion, the better they perform.

Output **"A is better"** when Video A performed better according to the evaluation rules.  
Output **"B is better"** when Video B performed better according to the evaluation rules.  
Output **"tie"** when both videos demonstrate similar capabilities.  
Output **"both are bad"** when both videos have hardly done anything related to the rules or have performed very poorly.

Before outputting the decision, you should list the relevant evidence from the videos to support your decision (within 80 words). Do not simply copy phrases from the rules.

You will make the decision across six major criteria:

1. Task Progress
2. Material Selection and Usage
3. Action Control
4. Error Recognition and Correction
5. Creative Attempts
6. Task Completion Efficiency

You should follow the output format below to organize your response:

---

Task Progress:
    - evidence: xxx  
    result: xxx

Action Control:
    - evidence: xxx  
    result: xxx

Error Recognition and Correction:
    - evidence: xxx  
    result: xxx

Creative Attempts:
    - evidence: xxx  
    result: xxx

Task Completion Efficiency:
    - evidence: xxx  
    result: xxx

Material Selection and Usage:
    - evidence: xxx  
    result: xxx

Overall results:
    - Task Progress: xxx  
    - Action Control: xxx  
    - Error Recognition and Correction: xxx  
    - Creative Attempts: xxx  
    - Task Completion Efficiency: xxx  
    - Material Selection and Usage: xxx

---

Note:
- If the evaluation rules include "e.g.", it is only an example, and you should not be limited to the listed examples. Consider all phenomena that conform to the major criteria.
- Task progress only considers the completion of key steps of the task and does not account for artistic qualities or similar aspects.
- In categories like task progress, action control, task completion efficiency, and material selection and usage, you should ideally choose either A or B as better.

\end{lstlisting}

\subsection{Prompt for Individual Video Rating}
\label{app:prompt_single_rating}
\begin{lstlisting}[language=, caption=Prompt for Individual Video Rating]
You are an expert of Minecraft and good at evaluating agents in the AI field.  
I will give you a task name, a grading criteria for this task, and a video of an agent performing the task.

The grading criteria has several major criteria (surrounded by ** **) and several evaluation rules under each major criterion.  
You need to score the agent's operations in the video based on the evaluation rules. The more the agent complies with the rules in the criteria, the higher the score it receives.  
If you think the agent's behavior does not relate to the stated rule, score 0.  
If you think the agent's behavior barely relates to the stated rule, score 0.1-0.3 
If the agent's behavior partially relates to the rules, score 0.4-0.6
If the agent's behavior is mostly related to the rules, score 0.7-0.9
If the agent's behavior is completely related to the rules, score 1.

If you believe the agent complies with the rule, you should list the relevant evidence from the video (within 50 words). Do not simply copy the phrases from the rules.  
Please assign an appropriate score across five dimensions, including task progress, material selection and usage, action control, error recognition and correction, creative attempts, and task completion efficiency, based on the final evidence.

You should follow the following output format to organize outputs. "xxx" is the placeholder. Evidence can be more than one.  
If there are multiple tasks, such as mining ore and crafting items, please provide a comprehensive evaluation, responding with only one overall score.

Output format:

Task Progress:
- evidence xxx
Score: xxx

Action Control:
- evidence xxx
Score: xxx

Error Recognition and Correction:
- evidence xxx
Score: xxx

Creative Attempts:
- evidence xxx
Score: xxx

Task Completion Efficiency:
- evidence xxx
Score: xxx

Material Selection and Usage:
- evidence xxx
Score: xxx

Overall Scores:
- Task Progress: xxx
- Action Control: xxx
- Error Recognition and Correction: xxx
- Creative Attempts: xxx
- Task Completion Efficiency: xxx
- Material Selection and Usage: xxx

Note:
- If the evaluation rules include "e.g.", it is only an example and you should not be limited to the listed "e.g." All phenomena that conform to the major criteria should be considered.  
- Task progress only considers the completion of key steps of the task and is unrelated to artistic qualities or other such aspects.  
- You should ignore the text shown on the video.  
- If the video has required materials for the task, they are automatically assigned by the system and cannot be counted in the task progress.   
- For combinations like "a and b or c," the average of the scores from tasks a and b should be calculated first, and then the higher value between this average and the score of task c should be taken as the final result.

\end{lstlisting}

\newpage

\subsection{Pseudo-Code Examples}

\begin{lstlisting}[language=Java, caption=Mineflayer Craft Task Pseudo-Code]
const doc = yaml.load(fs.readFileSync(task_conf, 'utf8'));  
// Extract the item name from the task description
const item_name = task_description.split('craft_a_')[1];   
// Execute each initialization command to set up the environment
doc.custom_init_commands.forEach(command => { 
    bot.chat(command); 
}); 
// Find the recipe for crafting the specified item
const recipe = bot.recipesFor(item_name, craftingTable);
// Attempt to craft the item
try {
    await bot.craft(recipe, count, craftingTablePosition);
    console.log(`${count} ${item_name} crafted successfully`);
} catch(err) {
    console.error('Failed to craft item:', err);
}
\end{lstlisting}

\begin{lstlisting}[language=Python, caption=MCU Evaluation Process Pseudo-Code]
from mcu_benchmark import MinecraftWrapper, VLM_Evaluator
from utility import load_config, check_success_and_save_video
from models import agent_creator

# Step 1: Load task configuration for the benchmark
config = load_config("build_house.yaml")  
# Step 2: Initialize the environment with MinecraftWrapper
env = MinecraftWrapper(config['env'], level=config['level'])
# Step 3: Initialize the agent (using custom model path and weights)
agent = agent_creator(model_path, weight_path).cuda()
agent.eval()  # Set the agent to evaluation mode
# Step 4: Get the initial state for the agent
state = agent.initial_state()
# Step 5: Start the environment and reset
obs, info = env.reset()
terminated, truncated = False, False
rollout_info = []
# Step 6: Agent's rollout
while not terminated and not truncated:
    # Get action from the agent and update state
    action, state = agent.get_action(obs, state)
    # Step the environment with the agent's action
    obs, terminated, truncated, info = env.step(action)
    # Save frames (visual feedback from the environment)
    rollout_info.append(info)
# Check if the agent succeeded in the task programmatically
success, video_path = check_success_and_save_video(rollout_info)
# Step 7: Evaluate the agent using a Vision-Language Model (VLM)
vlm_evaluator = VLM_Evaluator()
vlm_score = vlm_evaluator.evaluate(video_path, 'build_criteria.txt')
print(f"Success: {success}. VLM evaluation score: {vlm_score}")
\end{lstlisting}

\newpage

\subsection{Case study}

The following case clarifies the impact of each metric on evaluating generalization performance. Metrics such as task progress and material selection assess basic task alignment, while action control and task efficiency provide insights into optimization strategies.
Error correction and creative attempts, in contrast, measure higher-order generalization skills. These are critical for assessing agents in open-ended and complex scenarios, as they reveal resilience to failure and capacity for novel strategies.

While Video B outperformed Video A across most metrics, the weaknesses in creativity and error correction indicate areas where even high-performing agents fall short. Incorporating tailored training modules and broader tasks emphasizing these dimensions will enhance the benchmark’s utility for developing and evaluating generalist agents.

\begin{lstlisting}[language=, caption=Video Comparison Evaluation Results]
Task Progress:
- Video A: The agent collects dirt blocks and places them vertically but does not reach a reasonable height.
- Video B: The agent collects dirt blocks, places them vertically, and reaches a reasonable height.
result: B is better

Action Control:
- Video A: The agent places some blocks horizontally and in unrelated locations.
- Video B: The agent places blocks vertically without unnecessary actions
result: B is better

Error Recognition and Correction:
- Video A: The agent does not correct incorrectly placed blocks.
- Video B: The agent does not make any noticeable errors that need correction.
result: B is better

Creative Attempts:
- Video A: The agent does not show any creative attempts.
- Video B: The agent does not show any creative attempts.
result: tie

Task Completion Efficiency:
- Video A: The agent takes a longer time with unnecessary actions.
- Video B: The agent completes the task efficiently without unnecessary actions.
result: B is better

Material Selection and Usage:
- Video A: The agent uses dirt blocks but places some blocks horizontally and in unrelated locations.
- Video B: The agent exclusively uses dirt blocks and places them appropriately.
result: B is better

Overall results:
- Task Progress: B is better
- Action Control: B is better
- Error Recognition and Correction: B is better
- Creative Attempts: tie
- Task Completion Efficiency: B is better
- Material Selection and Usage: B is better
\end{lstlisting}

\newpage

\begin{lstlisting}[language=, caption=Individual Video Evaluation Results]
**Task Progress:**
- Evidence: The agent placed two snow blocks vertically and a carved pumpkin on top, but no Snow Golem was created.
- Score: Partially

**Action Control:**
- Evidence: The agent placed multiple unnecessary snow blocks around the structure.
- Score: Barely

**Error Recognition and Correction:**
- Evidence: The agent did not correct the placement of the carved pumpkin after failing to create a Snow Golem.
- Score: Barely

**Creative Attempts:**
- Evidence: No creative attempts or decorations observed.
- Score: None

**Task Completion Efficiency:**
- Evidence: The agent took excessive time with unnecessary placements and failed to complete the task.
- Score: Barely

**Material Selection and Usage:**
- Evidence: Correct materials (snow blocks and carved pumpkin) were used, but not effectively.
- Score: Partially

**Overall Scores:**
- Task Progress: Partially
- Action Control: Barely
- Error Recognition and Correction: Barely
- Creative Attempts: None
- Task Completion Efficiency: Barely
- Material Selection and Usage: Partially
\end{lstlisting}

\end{document}